\documentclass[3p]{elsarticle}

\usepackage{lineno,hyperref}

\usepackage{color,soul}

\usepackage{eurosym}

\usepackage{xcolor}

\usepackage{rotating} 

\usepackage{booktabs}

\usepackage{colortbl}

\usepackage{amsbsy}

\usepackage{amssymb}

\usepackage{url}

\usepackage{pifont}
\newcommand{\xmark}{\ding{55}}%

\usepackage{multirow}

\usepackage{textcomp} 

\modulolinenumbers[5]

\journal{Journal of \LaTeX\ Templates}

\begin{document}

\begin{frontmatter}

\title{{A Multi-Stage model based on YOLOv3 for defect detection in PV panels based on IR and Visible Imaging by Unmanned Aerial Vehicle}}



\author[flysight_address]{Antonio Di Tommaso\fnref{contrib}}
\ead{antonio.ditommaso@flysight.it}

\author[flysight_address]{Alessandro Betti\fnref{contrib}\corref{correspondingauthor}}
\ead{alessandrobetti53@yahoo.it}

\fntext[contrib]{These authors contributed equally to this work}

\cortext[correspondingauthor]{Corresponding author: Alessandro Betti}

\author[flysight_address]{Giacomo Fontanelli}
\ead{giacomo.fontanelli@flysight.it}

\author[flysight_address]{Benedetto Michelozzi}
\ead{benedetto.michelozzi@flysight.it}

\address[flysight_address]{FlySight S.r.l, Via Lampredi 45, 57121 Livorno, Italy}

\begin{abstract} 
As solar capacity installed worldwide continues to grow, there is an increasing awareness that advanced inspection systems are becoming of utmost importance to schedule smart interventions and minimize downtime likelihood.
In this work we propose a novel automatic multi-stage model to detect panel defects on aerial images captured by unmanned aerial vehicle by using the YOLOv3 network and Computer Vision techniques. The model combines detections of panels and defects to refine its accuracy and exhibits an average inference time per image of 0.98~s. The main novelties are represented by its versatility to process either thermographic or visible images and detect a large variety of defects, to prescript recommended actions to O\&M crew to give a more efficient data-driven maintenance strategy and its portability to both rooftop and ground-mounted PV systems and different panel types. 
The proposed model has been validated on two big PV plants in the south of Italy with an outstanding AP@0.5 exceeding 98\% for panel detection, a remarkable AP@0.4 (AP@0.5) of roughly 88.3\% (66.9\%) for hotspots by means of infrared termography and a mAP@0.5 of almost 70\% in the visible spectrum for detection of anomalies including panel shading induced by soiling and bird dropping, delamination, presence of puddles and raised rooftop panels. The model predicts also the severity of hotspot areas based on the estimated temperature gradients, as well as it computes the soiling coverage based on visual images. Finally an analysis of the influence of the different YOLOv3's output scales on the detection is discussed.
\end{abstract}

%


\end{frontmatter}


\section{Introduction}

\subsection{Motivation}
Photovoltaic (PV) installations continue to grow worldwide thanks to the increasing competitiveness of solar resource, the growing energy demand of developing countries and the effectiveness in addressing climate change and reducing global warming.
In the last decade it has been estimated that the overall installed PV capacity increased from about 40~GW in 2010 to 627~GW in 2019~\cite{Detollenaere 20}, and the investments in the sector are still growing. 

However, due to the exposure of outdoor PV panels and other plant components to environmental factors, they may experience premature thermo-mechanical stresses, which in turn can lead to a decrease of the PV plant efficiency and unexpected downtime depending on panel type. 
In particular, for common terrestrial temperatures it is estimated a 0.5\%/\textcelsius~ decrease of PV cell energy conversion efficiency with increasing temperature due to a reduction of the potential barrier of the PN junction~\cite{Libra 21}.
Crystalline PV modules are usually protected by an aluminium frame and a glass lamination in order not to be exposed directly to environmental agents. They are typically mounted on fixed structures oriented to the sun or on pan-tilt sun trackers to maximize their efficiency to the direct solar radiation. On the other hand, thin-film PV modules, whose response is more dependent to the diffuse solar radiation, are wrapped in a waterproof and flexible enclosure that is typically glued on fixed sloped or flat warehouse rooftops.  
Module degradation is roughly estimated in 1\% for crystalline PV modules~\cite{Chandel 15} and 3-4\% for thin-film~\cite{Suarez 19} on average per year due to intrinsic and extrinsic deficiencies.

The factors affecting defect occurrence are manifold~\cite{Kontges 14} and include, for example, crack (cell breakage, cracking of back sheet), cell oxidation or delamination, faults or disconnection of electrical components (e.g. junction box, by-pass diode), shading due to neighbour trees, buildings, soiling, bird dropping or snail tracks and rooftop slope.
A defect often appears in the form of a hotspot where the photovoltaic effect does not occur anymore, causing local overheating, leading in turn to destructive effects, such as cell or glass cracking, melting of solder, degradation of the solar cell~\cite{HSHP 21} or even fires and consequent irreparable catastrophes~\cite{Falvo 15}. 

Therefore, in order to keep high technical and economic performances of a PV system during its lifetime,  high-quality and cost-effective Operation \& Maintenance (O\&M) activities are sought.
Classical preventive or reactive maintenance strategies have been demonstrated to be sub-optimal due to the frequency of expensive visual on-site inspections or the cost induced by downtime and consequent component replacements, respectively.
As a consequence, in order to optimise simultaneously O\&M cost and plant efficiency, modern Data Analytics solutions based on predictive and early-detection strategies~\cite{Betti 21} are nowadays taking hold and need to be integrated in the remote monitoring platforms. 

Concerning panel inspection, traditional techniques include visual inspection and I-V curve tracing~\cite{Kontges 14}, which however are expensive, time-consuming, cause stops to energy production and are unfeasible for large PV plants or plants located in remote areas.
Alternative non-destructive techniques include instead Infrared (IR) and Electroluminescence (EL) Imaging~\cite{Kontges 14}: the former exploits the IR radiation emitted by the defect due to the local overheating, whereas the latter the disconnection of abnormal cell from the electric circuit, leading to a lack of  emitted IR radiation with respect to functional cells, when applying a current to the PV module~\cite{Petraglia 11}.
EL is able to recognize microcracks due to the high resolution, unlike IR imaging. However EL is an invasive technique conceived for analysis of a single PV module at a time, usually disconnected from the PV system and analysed indoor in a laboratory. 
IR imaging, especially if combined with aerial inspection by Unmanned Aerial Vehicles (UAVs)~\cite{Quater 14}, is instead a contact-less method that can be applied under real-time operating conditions directly on-site during the normal operation of the PV system. The method is particularly recommended for a fast inspection of hardly accessible PV plants, such as those installed on the building roof or facade. Further, it may be combined with advanced monitoring systems to identify the defect and estimate the related power loss~\cite{Libra 19}.
Latest aerial inspection best practices extend post-processing analyst cognition through the simultaneous acquisition of IR and VIS imageries~\cite{Zefri 18}.

All techniques, in particular those exploiting multiple sensors, produce a large amount of images which cannot be inspected visually by human operators because time-demanding, error-prone and expert-dependent: automated solutions for hotspot detections are therefore of utmost importance.
Furthermore, while IR imaging allows easily to detect generically a hotspot, further insights about the cause of the defect can be captured more easily only if monitoring also the Visible (VIS) spectrum.

\subsection{State of the art}
In the last few years, different studies exploited IR imaging to detect hotspots in PV systems. One of the early works concerning the applicability of IR imaging of PV modules under outdoor conditions has been proposed by Ref.~\cite{Buerhop 12}. 
Research works can be mainly grouped in three different categories depending on techniques involved and characterized by a progressively higher accuracy and lower processing time: (i) based on Image Processing, (ii) Machine Learning algorithms and (iii) deep Convolutional Neural Networks (CNNs).

Category (i) usually involves steps such as edge detection and Hough transform to detect the linear border of the PV module and application of pattern recognition methods based on the analysis of the intensity distribution of emitted radiation to detect faulty areas~\cite{Leotta 15}~\cite{Aghaei 15}.
In Ref.~\cite{Leotta 15} the model was first tested on images of indoor panels installed in laboratory and then on outdoor panels installed on-site, verifying the unsuitability for real-time operations. Ref.~\cite{Greco 17} proposed instead a double-stage procedure composed by a preliminary panel detection based on Hough transform and then hotspot detection based on a combination of a color based analysis with a model based one to rule out possible defect candidates due to heating localized at junction box. They obtain a remarkable computing speed of 25 FPS at a price of a F1-score not exceeding 60\%.
In general, however, the accuracy and processing rate of such methods (i) is unsatisfactory and unsuitable for implementation in O\&M services.

Group (ii) is instead usually based on two main steps: extraction of hand-crafted features and use as inputs for training a classification algorithm based on machine learning to discriminate modules as either nominal or defective. While such methods have usually higher accuracy than category (i), they require domain experience for feature extraction and may be unsuitable for real-time detection. 
In Ref.~\cite{Kurukuru 19} first-order and second-order texture features were extracted and used  to train a shallow NN to classify eight different fault classes, achieving finally a 91.7\% testing accuracy.
Ref.~\cite{Salazar 16} implemented a semi-automated process based on k-means clustering to determine the hotspot area and the corresponding average temperature.
Ref.~\cite{Deitsch 19} processed EL images of solar cells by extracting local descriptors at keypoints and inputting them into a Support Vector Machine (SVM) algorithm to classify the cells either as defective or functional. They also demonstrate that a more modern approach based on VGG19 network outperforms the SVM-based method. 

Methods (iii) are now progressively replacing previous techniques thanks to the superior performances of CNNs demonstrated in visual recognition tasks and the further advantage of automatic features extraction, which definitely leads to an increase of the computing speed. CNNs are often used in combination with IR images collected by UAVs, due to their recent diffusion.
Ref.~\cite{Pierdicca 18} uses a VGG16 network to classify PV cells either as damaged or functional achieving a F1-score up to roughly 70\%, and showing the impact of unbalancing on performances.
In Ref.~\cite{Oliveira 19} first a Gaussian filter is applied to reduce noise and a Laplacian operator to highlight edges. Then the image is segmented into defective and normal areas by using thresholding, and finally the binary mask is used to train a base network VGG16 to recognize disconnected substrings, hotspots and disconnected strings. 

Other works focused on visible images to detect the cause of hotspot, e.g. soiling~\cite{Hwang 20} or dust~\cite{Yap 15}\cite{Unluturk 19}. Authors in Ref.~\cite{Hwang 20} applied image processing techniques to UAV images to assess the soiling rate, while they postponed to a future work combination of artificial intelligence and statistical algorithms to evaluate the soiling distribution on PV panel surface. Ref.~\cite{Yap 15} and~\cite{Unluturk 19} simulated instead different dust coverage and demonstrated that image matching, or a shallow neural network fed with texture features extracted by gray level co-occurrence matrix, are effective solutions for evaluating the degree of pollution.

However, while the literature is quite rich, a limitation is that each panel is examined independently in laboratory~\cite{Oliveira 19}\cite{Yap 15}\cite{Unluturk 19}\cite{Mehta 18}\cite{Pierdicca 18}, requiring their disconnection from the field, stopping energy generation and making unsuitable their application on large-scale PV systems. This is also a simplified scenario with respect to UAV images captured on-site and differing for altitude, orientation, point of view, light intensity due to sun movement across the sky or cloud coverage, as well as presence of blurring due to a sudden camera movement caused by unexpected gusts of wind.

More critically, the interest is still in solving classification problems, whereas defect detection based on CNNs is still at an early stage. In fact, at the best of our knowledge, only few works have been presented so far~\cite{Mehta 18}\cite{Herraiz 20}\cite{Ashok 18}\cite{Pierdicca 20}\cite{Vlaminck 22}\cite{Bommes 21}.
In Ref.~\cite{Mehta 18} a four-stage process is proposed which completely avoids the time-demanding activity of manual labelling of training data. The model includes Mask FCNN to predict simultaneously the soiling localization mask and the percentage power loss, as well a webly supervised NN (WebNN) to predict the soiling category. It has been tested on a large dataset of more than 45k RGB images in the VIS spectrum of two different PV panels installed in laboratory, achieving a remarkable frame rate of 22 FPS.
Ref.~\cite{Pierdicca 20} proposed an anomaly cell detection systems based on instance segmentation of thermal images. They adopted the Mask R-CNN architecture~\cite{He 17} and benchmarked it with other image segmentation networks like U-Net, FPNet and LinkNet. Authors demonstrated that U-Net outperforms Mask-RCNN in terms of both accuracy and speed, but with the key-advantage of Mask R-CNN to directly returns the position of each single defective cell. 

In Ref.~\cite{Vlaminck 22} and~\cite{Bommes 21} a preliminary segmentation of aerial thermal image is performed based on Mask R-CNN to identify the singular PV module, then for each image patch anomaly detection is carried out in Ref.~\cite{Vlaminck 22} by using Faster R-CNN~\cite{Ren 15} based on the ResNet-50 backbone~\cite{He 16}, whereas in Ref.~\cite{Bommes 21} defective modules are classified by using ResNet-50 and providing also their exact location in the plant for fast on-site targeted repairs. The authors achieve a remarkable accuracy at the expense of a processing speed from almost 2~s to 4~s per singular PV module depending on plant~\cite{Bommes 21}.

In Ref.~\cite{Herraiz 20} a cascade three-stage detector is instead presented by using the double-stage region-based R-CNN. The IR image is initially rotated to align the panels, arrays of PV modules, to the image border by using the Sobel edge detector, then a panel detector is applied to detect panel areas and finally a hotspot detector identifies hotspots on such proposed regions. The model gains a sensitivity of almost 89\% for both panel and hotspot detection, exploiting also telemetry data to deliver the hotspot location with a mean error of almost 0.86~m. 
Finally Ref.~\cite{Ashok 18} discussed the applicability of YOLO~\cite{Redmon 16} on thermal images acquired by UAV for an unspecified collection of PV plants in India. While they demonstrated the suitability of the network for detection of hotspots of varying size, results were at an embryonic stage since no discussion was presented about performances and drawbacks of the proposed model. Furthermore, YOLO did not allow detection at multiple scales or recognition of crowded objects, issues overcome in later versions of YOLO.

\subsection{Paper contribution}
Urged by the aforementioned problems still unsolved, in this work we propose a novel multi-stage architecture for the detection of anomalies in images of PV panels collected on-site by UAV.
The model is composed by three main components: (i) a panel detector which detects the PV panel area, (ii) a defect detector which identifies the defects in the whole input image and (iii) a False Alarm Filter which removes false positives of defects detected outside the PV panel region proposals.

Unlike Ref.~\cite{Vlaminck 22} and~\cite{Bommes 21}, the proposed automated solution processes the full-sized image including many electrically interconnected PV modules, with the final objective to represent a satisfactory trade-off between accuracy and timing, consequently speeding up the maintenance inspection and optimizing the on-field interventions, which are especially critical for large PV systems.

While this work shares some common points with that presented by Ref.~\cite{Herraiz 20}, it extends and differs from the latter for the following reasons: (i) it operates on images of both the IR and visible spectrum, (ii) the application to two large PV plants of installed capacities of tens of MW and corresponding to different panel types (polycrystalline or thin-film) and installation types (ground or roof mounted), which is uncommon in Literature due to stringent data sharing policy, (iii) the simultaneous analysis of a large variety of defects, such as hotspot, the thermal stress induced by junction box, bird dropping, delamination and soiling, and including also issues not yet discussed adequately, such as raised rooftop panels and stagnant water (puddles), (iv) the prediction of the hotspot severity and the prescription of the related maintenance action, (v) the estimation of the panel area affected by soiling occurrence, which is an important feedback for O\&M operator to quantify the amount of PV cells affected by power loss and schedule more effective maintenance operations, (vi) the application of the end-to-end single-stage detector YOLOv3~\cite{Redmon 18} as core network of the multi-stage model and, finally, (vii) the portability of the model architecture on plants of different size, locations and panel types and for images collected in different wavelength bands, once a sufficient training statistics is available. 


\section{Case study}
\label{case_study}

\begin{figure}[ht]
	\centering
	\includegraphics[width=0.5\textwidth]{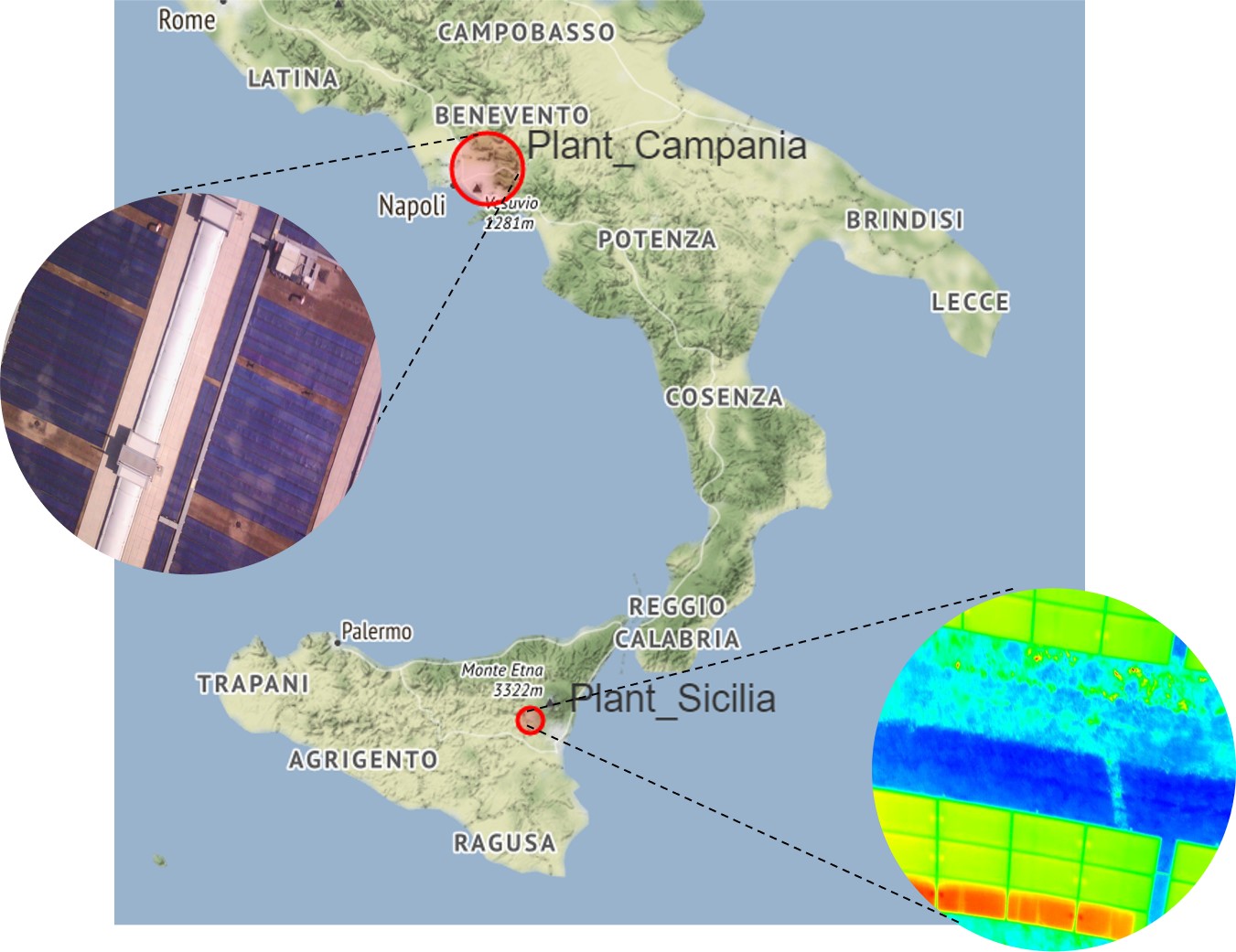}
	\caption{Location of the two considered PV plants in Italy. The red circle size is proportional to the installed capacity. The insets show an example of images captured by UAV.
	}
	\label{fig:plants_location}
\end{figure}

Throughout the paper, we will refer to two PV plants, here below denominated as Plant\textunderscore Sicilia and Plant\textunderscore Campania, which are located in the southern of Italy, as shown in Fig.~\ref{fig:plants_location}. 

\subsection{PV plants details}
Plant\textunderscore Sicilia  is composed by polycrystalline modules installed on the ground and is situated in Sicily, a region in the South of Italy. The photovoltaic installation has a nominal capacity of 9 MW and consists of over 20 thousand panels, where each panel is a PV array collection of individual PV modules electrically connected together. It generates over 7.2 million kilowatt-hours per year, enough to meet the energy needs of almost 2000 households, as well as avoiding the atmospheric emission of over 3800 tonnes of CO\textsubscript2 per year.

Plant\textunderscore Campania is instead composed by thin-film photovoltaic modules installed on the flat roof of 56 commercial and logistics buildings in the italian region of Campania.
The plant has a nominal capacity of 21 MW and can produce approximately 33 million kilowatt-hours of power each year, satisfying the consumption needs of 13,000 households and avoiding the atmospheric emission of more than 21 thousand tonnes of CO\textsubscript2 per year. 

\subsection{Datasets} \label{Datasets}
The inspection was performed by means of a drone Sigma Ingegneria Efesto MKII equipped with a flight controller DJI A2 and a gimbal system designed by Sigma Ingegneria. 
Two cameras were mounted on the drone: Workswell WIRIS 640 second Gen and MAPIR Survey3N RGB. 
MAPIR camera was used to get high resolution images in the visible spectrum (VIS-HR), whereas WIRIS camera took thermal IR images (LWIR) and aligned low resolution visible images (VIS-LR). 

\begin{figure}[!tbh]
\centering
	\vspace{0cm}
	\includegraphics[width= \linewidth, keepaspectratio, trim={1cm 0 0 0}]{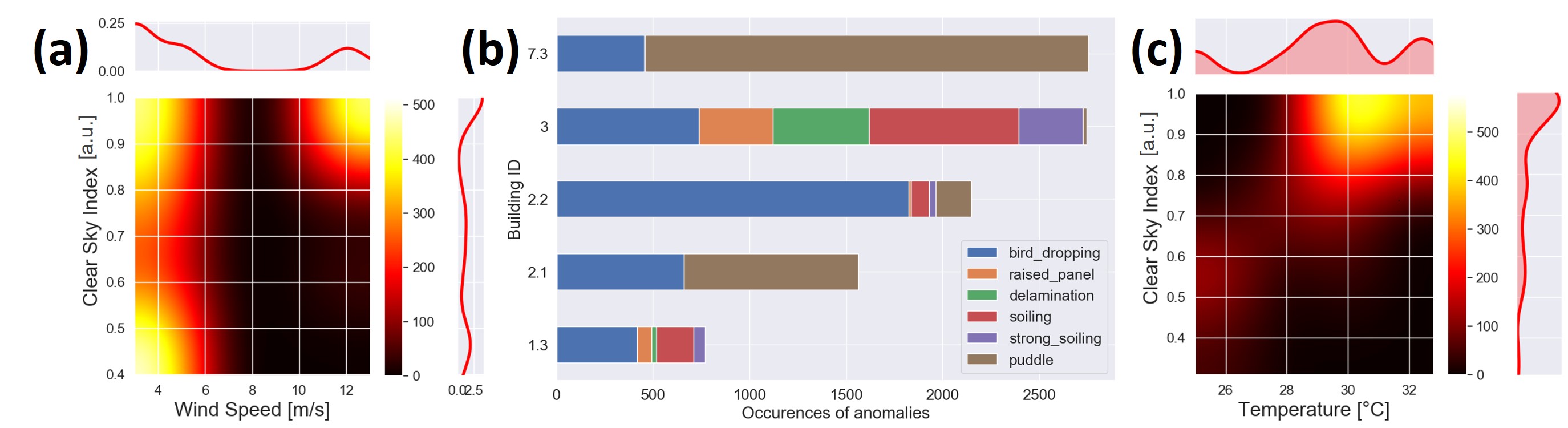}
	\caption{2D distribution of aerial images in the plane (a) (wind speed, CSI) for Plant\_Sicilia and (c) (ambient temperature, CSI) for Plant\_Campania, respectively, where CSI is the ratio between the measured global horizontal irradiance and the clear-sky horizontal irradiance. (b) Number of occurrences of defects for Plant\_Campania grouped by building and defect class.}
	\label{fig:data_features}
\end{figure}

In general, the representativeness of the collected data, as well as the reliability of results achieved on such sample, depend on two main points: the complexity of the scenario under investigation and the complexity of the learning algorithm, which will be discussed in the next sections. 
Concerning the first point, to make dataset really representative of the two PV systems, the inspections for each plant were conducted in different days and times in order to collect a rich statistics in terms of illumination conditions, background and weather. 
In particular, data acquisition on Plant\_Sicilia was performed in the spring 2018 in sunny or partial cloudy days with ambient temperature higher than 15~\textcelsius, ~days experiencing variable wind speed and direction (often from the south), and Clear Sky Index (CSI), which is the  ratio between the global horizontal irradiance and the global horizontal irradiance in clear sky conditions, ranging from 0.4 to 1 (Fig.~\ref{fig:data_features}(a)). In particular, Sicily is sensitive to the “sirocco”, the hot wind coming from Africa, which can bring temperatures even higher than 20~\textcelsius ~in winter, and at 40~\textcelsius ~in summer.

Similarly, inspections on Plant\_Campania were realized in summer 2018 with air temperature higher than 26~\textcelsius ~and usually in condition of high solar irradiation (Fig.~\ref{fig:data_features}(c)). Also, flights were scheduled on different buildings characterized by various levels of scene brightness, soiling and bird dropping severity, dust deposition due to dirty rainfall, stagnant water, as well as panel oxidization due to damaged adhesion to the building roof (Fig.~\ref{fig:data_features}(b)).

According to standard Machine Learning methodology, each dataset has been split randomly in training (70\%), validation (15\%) and test (15\%) by implementing stratified sampling for each defect class. In particular, the validation set has been used for model optimization and performance evaluation, whereas the test set for performance evaluation only. 
More details are provided in the following subsections.

\subsubsection{Plant\textunderscore Sicilia's dataset}
A nadir thermal camera mounted on the bottom of the drone captures a LWIR image containing PV panels and context and saves a 16-bit gray-scale tiff image containing the raw readings of the sensor.
The raw image is then post-processed to get the temperature in the scene by using a deterministic algorithm, based on Plank’s and Stephan Boltzmann laws, and considering the emissivity of the object being inspected. The estimated temperature values for each pixel are then normalized according to an 8-bit per pixel format and min-max normalization, and the obtained gray-scale matrix is finally converted into a false-color RGB image based on a rainbow color palette. 
As a consequence, the false-color RGB image is based on an adaptive scale that takes into account the coldest and warmest objects present in the scene.

\begin{figure}[!tb]
\centering
	\vspace{0cm}
	\includegraphics[width=11cm, trim={2cm 0 2cm 0}, keepaspectratio]{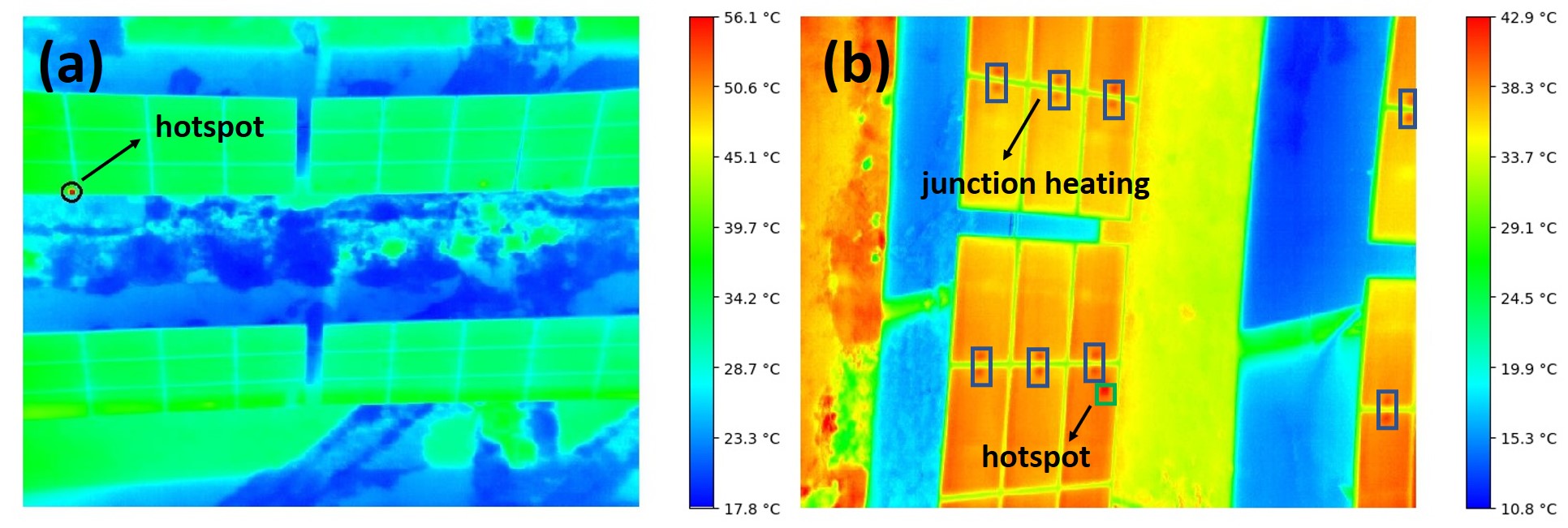}
	\vspace{0cm}	
	\caption{(a) Hotspot example highlighted by a black circle; (b): examples of hotspot (green rectangle) and junction box (blue rectangle) instances. The colorbar denotes the scene temperature expressed in units of Celsius degree. The PV system under consideration is Plant\textunderscore Sicilia.}
	\label{fig:defect_example_sicilia}
\end{figure}

Six flights were executed collecting around 500 RGB images for each. 
In particular, we selected and manually annotated 2038 radiometric LWIR images all of size 640x512 for hotspot detection corresponding to different UAV flights and including PV panels with different shape, size, and orientations.
Typically, at diurnal hours when UAV flights are performed, PV modules are often 15-20~\textcelsius ~warmer than ground, which is always present in the background of the image (Fig.~\ref{fig:defect_example_sicilia}). Hotspots are even hotter and appear often as a red region over a cold blue background, because the defective cells are negatively biased and photovoltaic effect is replaced by heat energy dissipation. In addition, local heating with respect to neighborhood normal modules may vary from few Celsius degrees (Fig.~\ref{fig:defect_example_sicilia}(b)) to tens of Celsius degrees (Fig.~\ref{fig:defect_example_sicilia}(a)).

Since module temperature depends on environmental factors such as ambient temperature, irradiance and presence of wind, as the latter change, module temperature and its efficiency will also vary, with usually a temperature rise above air and ground temperatures during the day when the sun shines on the PV modules. The use of an adaptive scale for the UAV mounted camera, as explained above, and collected images including simultaneously both panels and ground, allows therefore to efficiently apply the AI-based model to scenarios corresponding to different environmental conditions.

Due to the presence of frequent local overheating in correspondence of junction boxes, as discussed for example also in Ref.~\cite{Salazar 16}, we annotated two different classes: (i) hotspot and (ii) thermal stress induced by junction box.
Fig.~\ref{fig:defect_example_sicilia}(b) shows examples of Ground Truth Boxes (GTBs) for both classes: class (ii) appears as a couple of hot points which, without a proper annotation, may be misclassified as hotspot by the detector. Concerning class (i), we did not distinguish between heating affecting one cell or group of contiguous cells, as instead in Ref.~\cite{Ashok 18}, since our statistics is mainly limited to single defective cells. 

Tab.~\ref{tab:statistics_sicilia_dataset} shows the statistics available for training, validation and test sets. The dataset includes 1426 training images and 306 images for both validation and test. 
As may be seen in Tab.~\ref{tab:statistics_sicilia_dataset}, a severe class unbalancing occurs, being the minority class "hotspot" consisting of only 5.44\% samples of the overall dataset, i.e. almost 17 times smaller than the "junction" class. The statistics includes also around 6k annotated panels, with 926 samples left out for test.

\begin{table}[h]
\small
\centering
	\begin{tabular}{c|c|c|c}
	\rowcolor[gray]{0.9} \textbf{Dataset} & \textbf{hotspot} & \textbf{junctions} & \textbf{panels} \\
	\hline                       
	Train & 341 (70.75\%) & 5920 (70.62\%) & 4152 (69.42\%)\\
	Validation & 70 (14.52\%) & 1168 (13.93\%) & 903 (15.10\%)\\
	Test & 71 (14.73\%) & 1295 (15.45\%) & 926 (15.48\%)\\
	\hline
	\rowcolor[rgb]{ .886,  .937,  .855} Overall & 482 & 8383 & 5981\\
	\hline 
	\end{tabular} 
	\caption{\label{tab:statistics_sicilia_dataset}Dataset statistics for Plant\textunderscore Sicilia dataset. The number of GTBs for each class of defect and panel detectors is reported for the overall, training, validation and test sets. The percentage of the overall dataset is also shown in round brackets.}
\end{table}

\begin{figure}[!tb]
\centering
	\vspace{0cm}
	\includegraphics[width=11cm, keepaspectratio]{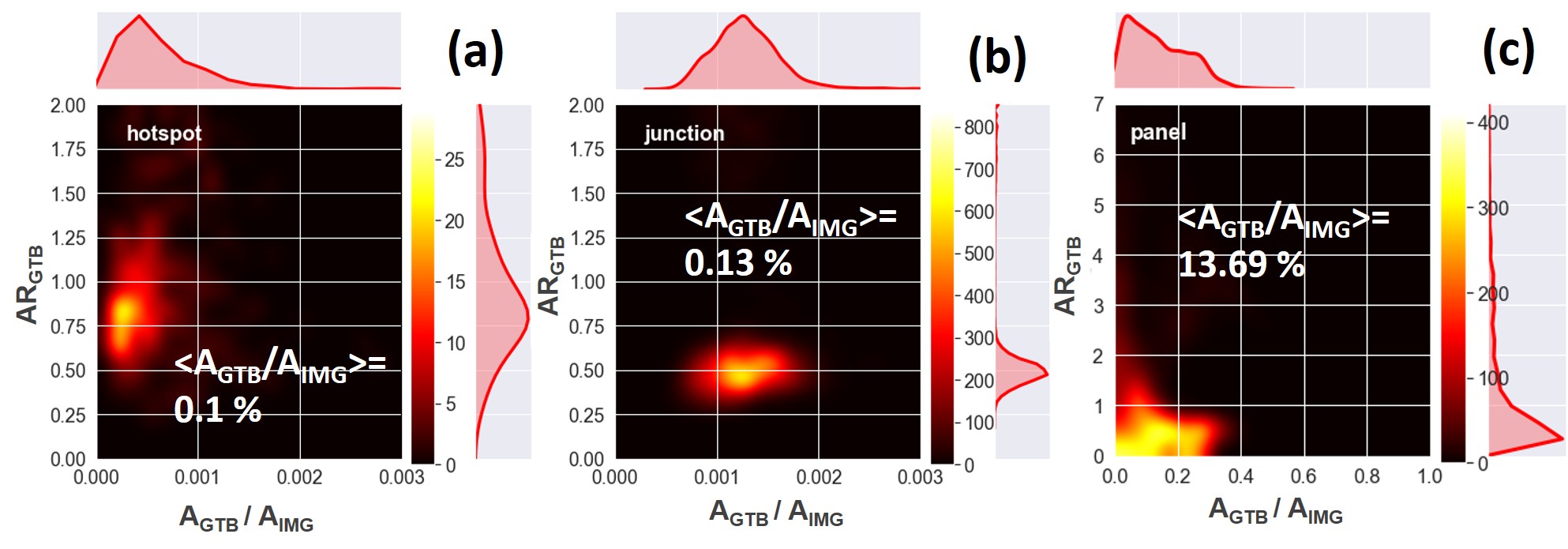}
	\vspace{0cm}
	\caption{2D distribution of GTBs of (a) hotspots (b) junction boxes and (c) PV panels in the plane (A\textsubscript{GTB}/A\textsubscript{IMG}, AR\textsubscript{GTB}), where A\textsubscript{GTB} is the GTB area normalized with respect to the image area A\textsubscript{IMG} and AR\textsubscript{GTB} is the GTB aspect ratio, i.e. the ratio between GTB width and height. The marginal distributions are also reported as shaded red areas. The plant under consideration is Plant\textunderscore Sicilia.} 
	\label{fig:eda_sicilia}
\end{figure}

Figure~\ref{fig:eda_sicilia} gives an overview of the main features of the proposed dataset.
Hotspots are challenging targets with a tiny average area of just 0.1\% of the whole image area (Fig.~\ref{fig:eda_sicilia}(a)). In addition, junction box area is almost comparable to hotspots (Fig.~\ref{fig:eda_sicilia}(b)), therefore enhancing the likelihood of false positives.
The GTB of PV panels have roughly an area smaller than 40\% of the whole image area and an aspect ratio peaked around 0.5, i.e. it is more likely to encounter panels oriented vertically than horizontally (Fig.~\ref{fig:eda_sicilia}(c)).
The Probability Density Function (PDF) of the panel area is peaked around 5\% of the whole image area and almost flat between 20\% and 30\%, with a mean area of about 13.7\% (Fig.~\ref{fig:eda_sicilia}(c)).

\subsubsection{Plant\textunderscore Campania's dataset}

We prepared and manually annotated 1500 digital VIS-LR images of size 1600x1200 in order to identify defects and related cause, thanks also to a deeper inspection on VIS-HR images. The images were captured over the roof of five different buildings in order to increase statistics and model robustness.
Six different defect classes were annotated: (i) puddle, (ii) soiling, (iii) strong soiling, (iv) delamination, (v) raised panel and (vi) bird dropping.  

More in details, "puddle" corresponds to the accumulation of water on the panels installed on flat rooftops. 
For sloped rooftops or crystalline panels mounted on trackers it has been verified in Ref.~\cite{Gaur 14} that water flow increases the module efficiency by decreasing the module temperature and acting as a cleaner. On the other hand, on flat rooftops, not properly working drainpipes or rooftop hollows produce puddles of stagnant water that will persist until complete evaporation. In case of a dirty rainfall, suspended dust will leave a soiling layer that will grow in time. As a consequence, when puddle cover PV panels they are eligible to become strong soiling, progressively altering the module operation and decreasing the irradiance absorbed by the module.
Even more dangerous, if water accumulation persists in time, it may penetrate into the panel through eventual micro-cracks or delamination points, causing corrosion, leakage current increment and eventual short-circuits. It can be also an early symptom of sinking for roof-mounted PV modules. Therefore maintenance interventions are recommended to identify and remove such potential issue. 

Classes "soiling" and "strong soiling" are instead related to deposition of dust, drifting sand, car and industrial exhaust fumes, etc. on the PV panels, causing partial shadowing and greatly reducing the absorption of solar irradiance. They appear usually as spots with different and irregular size. In particular, we label as strong\_soiling instances characterized by a dense pattern after a visual inspection, unlike class soiling for which deposition is more sparse.

"Delamination" occurs when the adhesion between glass, encapsulant, solar cells and back layers is compromised because of contamination (i.e. improper cleaning of the glass) or environmental factors~\cite{Kontges 14}. It is usually followed by moisture ingress and corrosion (panel oxidation), causing optical reflection and subsequent decrease in module efficiency. When the delamination takes place, it can affect a single cell or multiple cells. In any case, it is required the replacement of the whole PV panel. 

Solar cells in Plant\_Campania are deposited as a thin layer on the flexible backing strip having a coating of adhesive applied that adheres directly to the roof surface. Flexible solar strips are particularly suitable for large coverages with any kind of tilt angle such as industrial floor, business and sports centres. They have typically a lower efficiency than mono or polycristalline modules but a higher sensitivity to diffuse radiation, lower weight and simplify installation. However it may happen that one or more pieces of panel unglued from the base: we call it as "raised panel". 
The issue typically insists in the junction area between two consecutive module cells, where the panel thickness is lower due to the absence of amourphous silicon. The reason of such behaviour is not really resolved yet: possible causes include heat dissipation, the flexibility of the panel itself and the erroneous glue application during installation. Raised panels need to be glued again to avoid integrity damages and its related consequences. The latter may consist in delamination, lower insulation resistance and correspondent increase of the leakage voltage values, which in turn may invalidate the wet insulation and lead to PV plant interruption.

Finally class "bird dropping" corresponds to the accumulation of bird dropping on the panel. It is one of the most aggressive type of soiling as it can burn into the glass and cause hotspots under intense sunlight conditions. This defect usually occurs in multiple instances distributed over the same panel.

\begin{figure}[!tb]
\centering
	\vspace{0cm}
	\includegraphics[width=11cm, keepaspectratio]{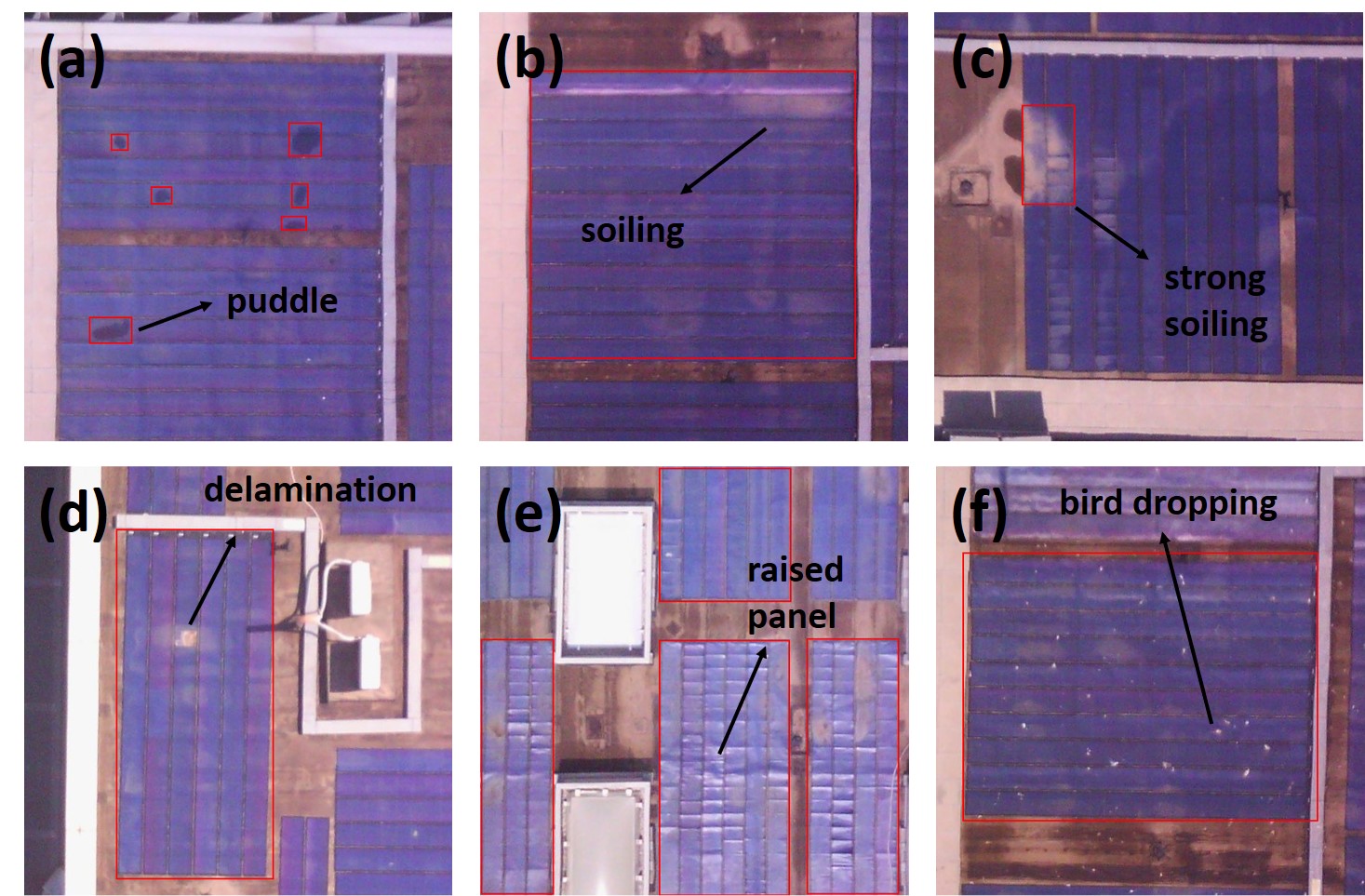}
	\vspace{0cm}	
	\caption{Examples of GTBs (red rectangles) for the defect classes of Plant\textunderscore Campania: (a) puddle, (b) soiling, (c) strong soiling, (d) delamination, (e) raised panel and (f) bird dropping.}
	\label{fig:defect_example_campania}
\end{figure}

Fig.~\ref{fig:defect_example_campania} shows some examples of GTBs for the mentioned classes. As can be seen, the defects were annotated either at instance (puddle, strong soiling) or panel (soiling, delamination, raised panel, bird dropping) level depending on the localization of their features over well-defined and delimited areas or their irregular distribution over the whole PV panel, respectively.

\begin{figure}[!ht]
\centering
	\vspace{0cm}
	\includegraphics[width=12cm, keepaspectratio]{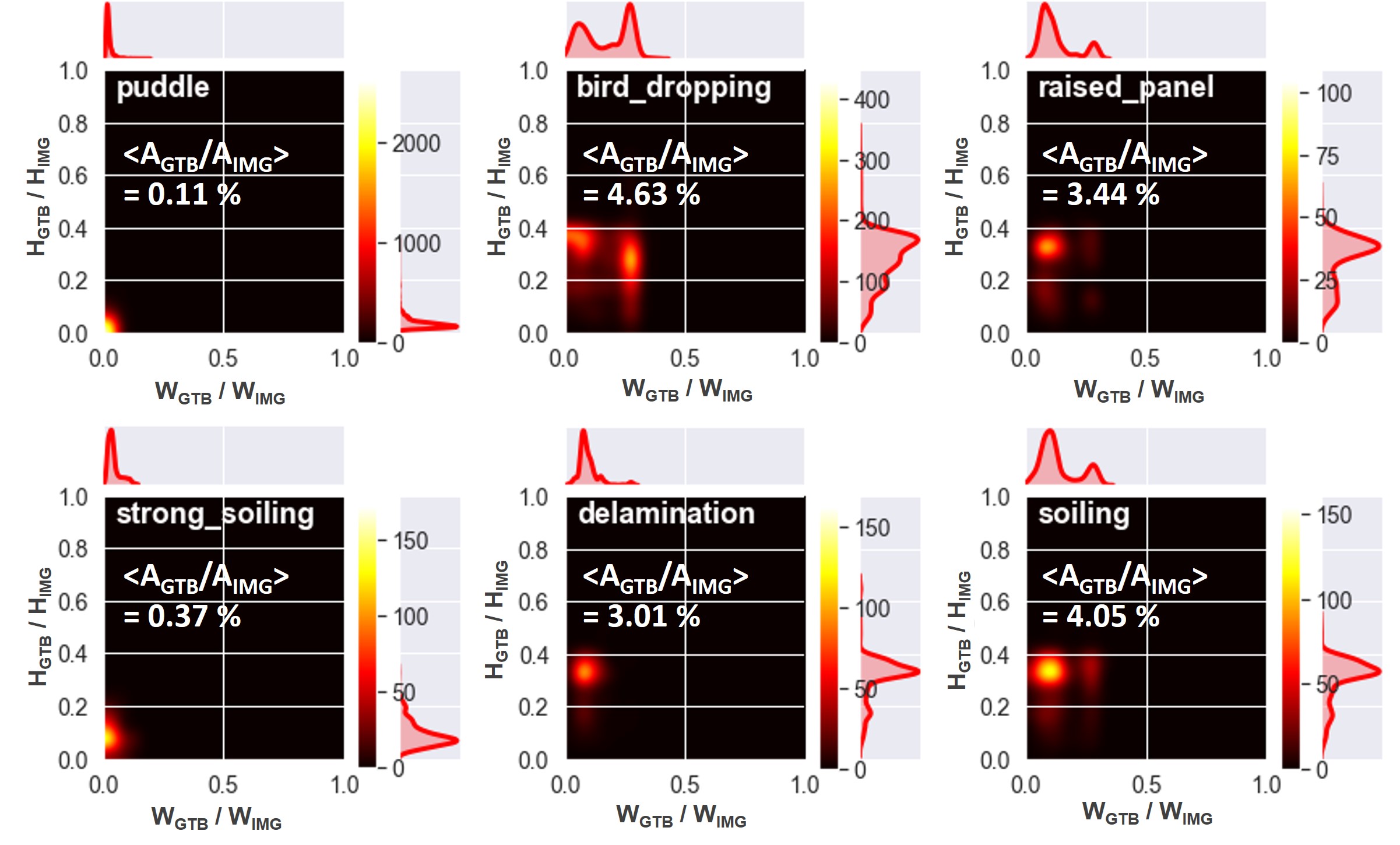}
	\vspace{0cm}	

	\caption{2D distribution of GTBs in the plane (W\textsubscript{GTB}/W\textsubscript{IMG}, H\textsubscript{GTB}/H\textsubscript{IMG}) for each defect class, where W\textsubscript{GTB} (H\textsubscript{GTB}) is the GTB width (height) and W\textsubscript{IMG} (H\textsubscript{IMG}) is the image width (height). The marginal distributions are also shown as shaded red areas. The PV system under consideration is Plant\textunderscore Campania.} 
	\label{fig:eda_defects_2D_campania}
\end{figure}

Tab.~\ref{tab:statistics_campania_dataset_class} reports the statistics available for training, validation and test sets for defects and panels, as well as the statistics singularly for each defect class. 
The dataset includes 1050 images for training and 225 for both validation and test. It contains, globally, 9984 defect instances grouped in 6 classes, as shown in Tab.~\ref{tab:statistics_campania_dataset_class}. The majority (minority) class is represented by bird dropping (strong soiling) with roughly 41.0\% (4.3\%) of the statistics available.
As may be seen in Fig.~\ref{fig:eda_defects_2D_campania}, the defect size can be clustered in two main groups depending on the adopted annotation either at instance or panel level, with an average dimension of the order 0.1-0.4\% in the first case, i.e. for classes puddle and strong soiling, and roughly one order of  magnitude higher for the remaining classes. The most challenging class to detect is represented by puddle, with a 95\% quantile of the GTB area of about 0.4\% of the image dimension. 


\begin{table}[h]
\small
\centering
\begin{tabular}{c|c|c|c|c}
\hline
\rowcolor[gray]{0.9} \textbf{class} & \textbf{Train} & \textbf{Validation} & \textbf{Test} & \textbf{Overall}\\
\hline
strong\textunderscore soiling & 270 (3.96\%) & 72 (4.32\%) & 86 (5.74\%) & 428 (4.29\%)\\
raised\textunderscore panel & 311 (4.56\%) & 72 (4.32\%) & 92 (6.14\%) & 475 (4.76\%)\\
delamination & 347 (5.09\%) & 103 (6.18\%) & 73 (4.87\%) & 525 (5.26\%)\\
soiling & 675 (9.90\%) & 197 (11.81\%) & 188 (12.55\%) & 1060 (10.61\%)\\
puddle & 2348 (34.44\%) & 541 (32.43\%) & 514 (34.31\%) & 3403 (34.08\%)\\
bird\textunderscore dropping & 2867 (42.05\%) & 683 (40.95\%) & 545 (36.38\%) & 4095 (41.01\%)\\
\hline
\rowcolor[rgb]{ .886,  .937,  .855} \textbf{Defects} & 6818 & 1668 & 1498 & 9984 \\
\hline
\rowcolor[rgb]{0.88,1,1} \textbf{Panels} & 12718 & 2775 & 2687 & 18180\\
\hline
\end{tabular}
\caption{\label{tab:statistics_campania_dataset_class} Number of defect instances grouped by class for Plant\textunderscore Campania. In the brackets the percentage with respect to the corresponding dataset is also shown. The last two rows indicate the overall number of defect and panel instances available.}
\end{table}

\begin{figure}[!tb]
\centering
	\vspace{0cm}
	\includegraphics[width=8cm, keepaspectratio]{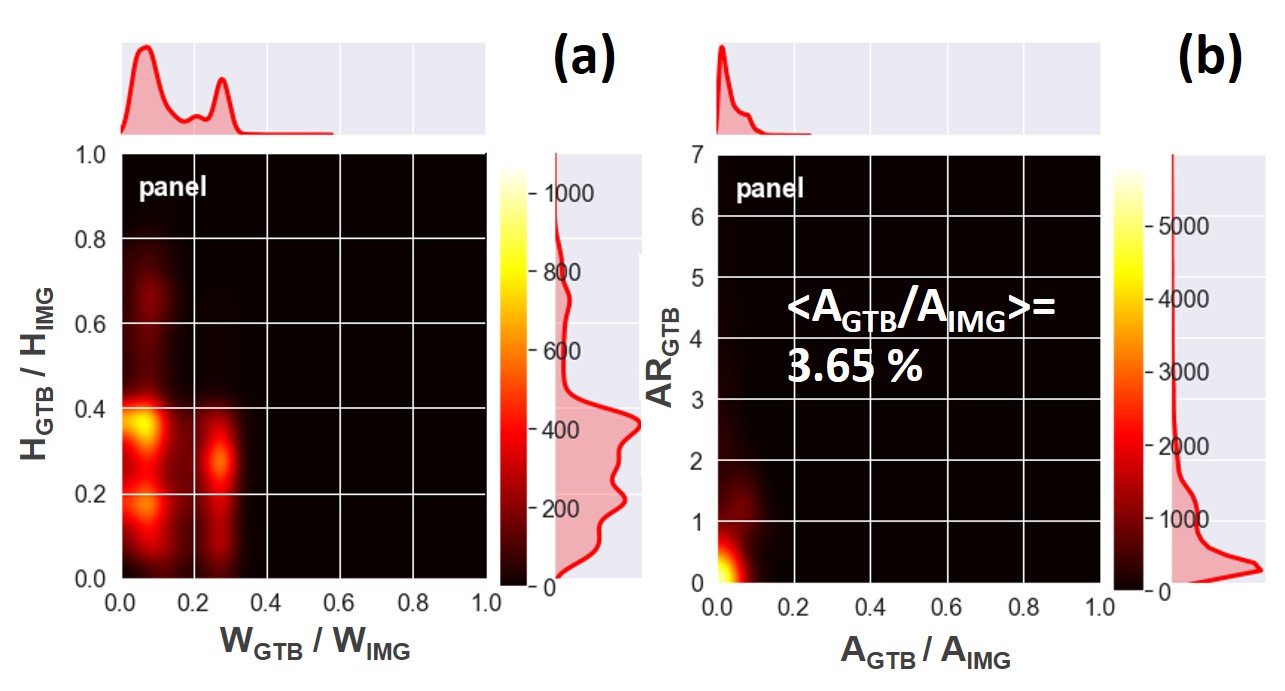}
	\vspace{0cm}	
	\caption{2D distribution of GTBs of PV panels in the plane (a) (W\textsubscript{GTB}/W\textsubscript{IMG}, H\textsubscript{GTB}/H\textsubscript{IMG}) and (b) (A\textsubscript{GTB}/A\textsubscript{IMG}, AR\textsubscript{GTB}) where W\textsubscript{GTB} (H\textsubscript{GTB}) is the GTB width (height), W\textsubscript{IMG} (H\textsubscript{IMG}) is the image width (height), A\textsubscript{GTB} (A\textsubscript{IMG}) is the GTB (image) area and AR\textsubscript{GTB} is the GTB aspect ratio. The marginal distributions are also shown as shaded red areas. The PV system under consideration is Plant\textunderscore Campania.}
	\label{fig:eda_panels_campania}
\end{figure}

Concerning panels, almost 18k instances were annotated with 2687 samples left out for test (Tab.~\ref{tab:statistics_campania_dataset_class}). In addition, they exhibit a side length usually not exceeding the 40\% of the image dimension (Fig.~\ref{fig:eda_panels_campania}(a)), with a mean area of almost 3.6\% and an aspect ratio close to 0.5 (Fig.~\ref{fig:eda_panels_campania}(b)).


\section{Methods} \label{Methods}
The proposed approach consists of a multi-stage architecture composed by three main processing modules and may be easily applied to aerial images in both the IR and VIS spectrum with modest customization: (i) a Panel Detector detecting the PV panel areas, (ii) a Defect Detector, which identifies the defect instances by processing the whole input image, and (iii) a False Alarm Filter, which finally post-processes the outcomes of the previous modules and filters out the defect proposals outside the predicted panel areas.

\begin{figure}[!tb]
\centering
	\vspace{0cm}
	\includegraphics[width=\linewidth, keepaspectratio]{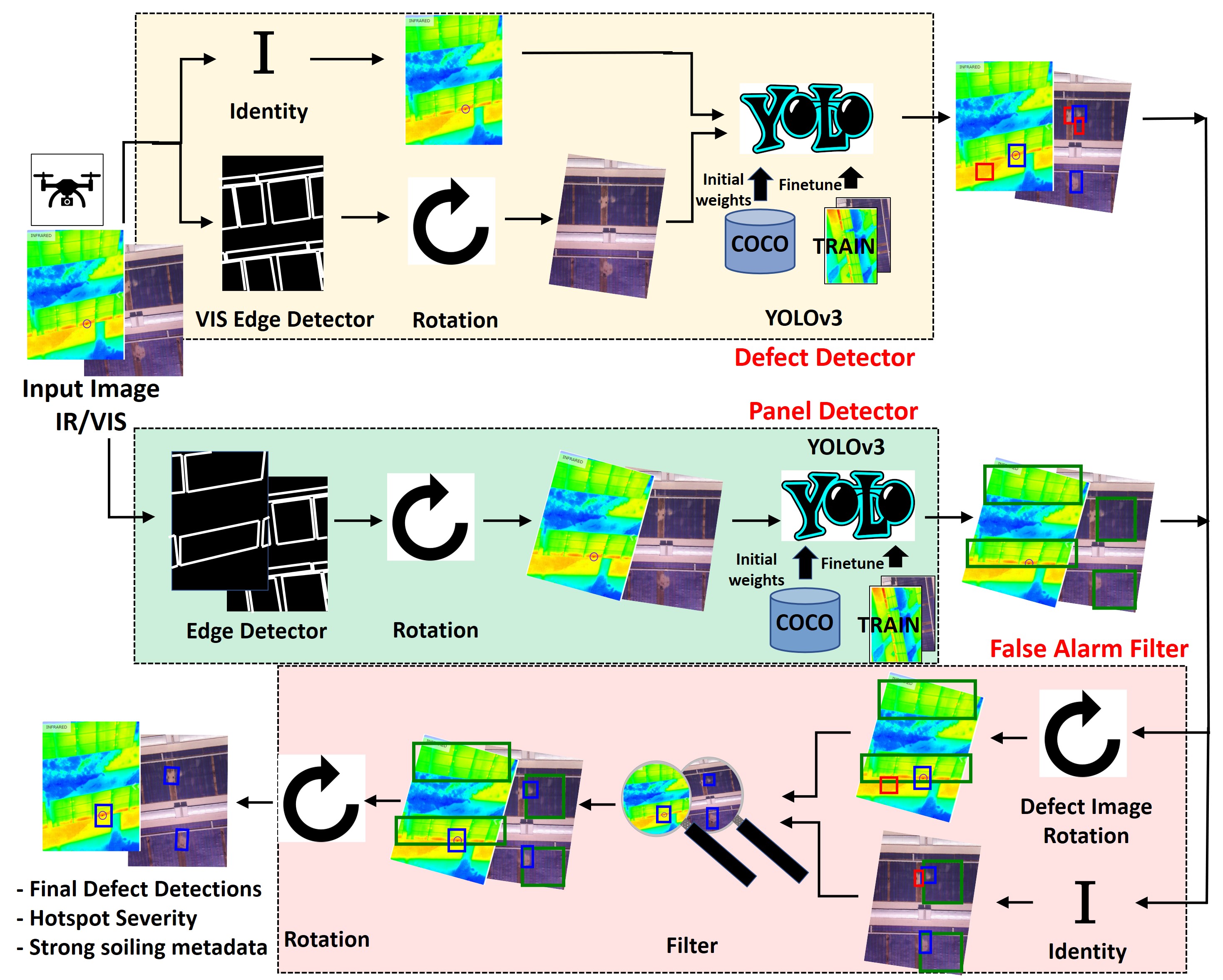}
	\vspace{0cm}	
	\caption{Pipeline of the proposed model. The input image (either IR or VIS) is fed into the Panel and Defect detectors based on YOLOv3 architecture and Computer Vision. The resulting outcomes are processed by the False Alarm Filter, which screened out all defect proposals (red rectangles) outside the detected panel areas (green rectangles), returning only defects inside (blue rectangles) expressed into the reference system of the input image. In addition, in case of IR images the defect severity and the corresponding suggested maintenance action are provided, whereas for VIS images the strong\_soiling coverage is returned.} 
	\label{fig:model_pipeline}
\end{figure}

\begin{figure}[!tb]
\centering
	\vspace{0cm}
	\includegraphics[width=\linewidth, keepaspectratio]{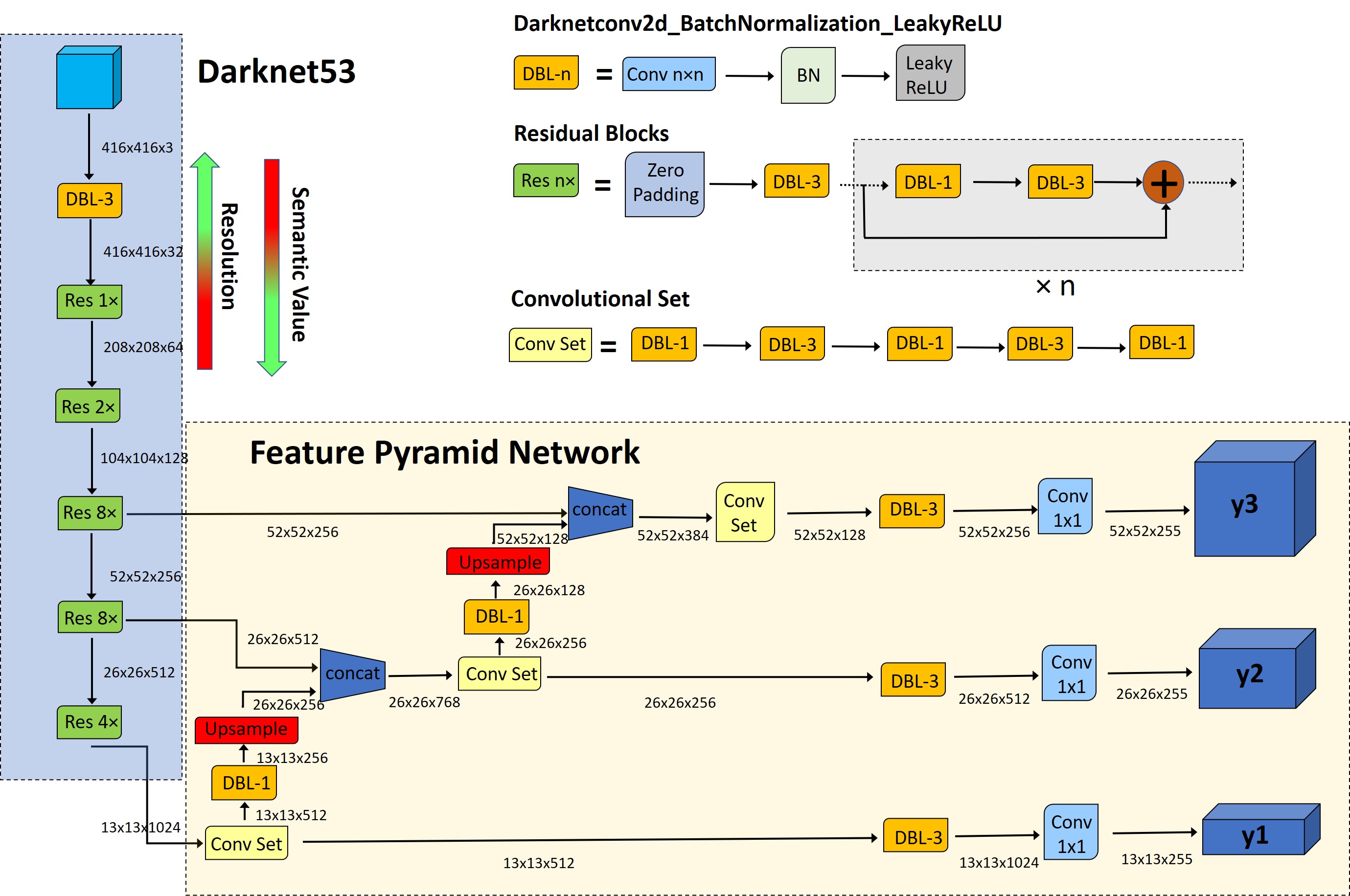}
	\vspace{0cm}	
	\caption{YOLOv3 architecture composed by the backbone Darknet-53 and the Feature Pyramid Network making multi-scale object prediction. $\bigoplus$ denotes the element-wise summation operation, whereas "concat" corresponds to lateral connection along depth.  In the figure we assume C= 80 classes (COCO dataset~\cite{Lin 14}).}
	\label{fig:yolov3}
\end{figure}

The model pipeline is depicted in Fig.~\ref{fig:model_pipeline}: the input image captured by UAV is fed into the Panel and Defect detectors, both based on a sequence of Computer Vision techniques, geometrical transformations and Artificial Intelligence based on YOLOv3 deep neural network (Fig.~\ref{fig:yolov3}). 
Since YOLOv3 works with rectangular bounding boxes, in case of ground truth delimiting the whole panel area, performances are better if the edges of the panels are aligned to the edges of the image.
Consequently, for panel detection, and also for defect detection in VIS images where ground truths for soiling, delamination, raised panels and bird dropping extend over the whole panel rectangles (Fig.~\ref{fig:defect_example_campania}), the image is preliminary rotated to maximize localization skills by first detecting the linear edges of the panels and then rotating it according to the identified predominant directions. 
Then defects and panels are detected on the resulting images by using YOLOv3.

Finally the outcomes are ingested into the False Alarm Filter which operates as following: (i) it expresses defect proposals into the reference system of panels (actually the rotation is necessary only for IR images), (ii) it discards the defect proposals (red rectangles) detected outside the proposed panel areas (green rectangles) which can be due to sun's glare or other external agents and (iii) it returns only the defects identified inside the panels (blue rectangles) by anti-transforming them into the original reference system of the input image. 
Details about the building blocks of the model are provided in the following sections.

\subsection{YOLOv3 network}
YOLOv3~\cite{Redmon 18} is a single-stage end-to-end network composed by a backbone and a head subnet (Fig.~\ref{fig:yolov3}). The backbone is represented by the Darknet-53 network and is responsible for features extraction from input image. In order to expand the receptive field and get more contextual information useful for detecting small targets, it applies downsampling based on strided convolutions. To overcome the problem of vanishing gradients with the increase in network depth and extract more robust features, it implements skip connection by using a sequence of residual units~\cite{He 15}.

The head subnet is built on top of the backbone and makes prediction at three different scales 13$\times$13, 26$\times$26 and 52$\times$52 by means of a Feature Pyramid Network (FPN)~\cite{Lin 16}.
FPN alleviates the problem of detecting objects of different scales by implementing feature fusion via a top-down pathway and lateral connections between feature maps of different resolutions.
Indeed deep low-resolution features bring a high semantic value, but the position information of the corresponding feature maps is weakened, resulting more suitable for detecting large objects due to the large receptive field. On the other hand, earlier high-resolution features have semantically low value and are better for detection of smaller objects due to limited receptive field.
FPN makes YOLOv3 more accurate than Single Shot Detector~\cite{Liu 16}, where detection is done separately on feature maps having different scales, and competitive with double-stage detectors, such as Faster R-CNN~\cite{Ren 15}, which are however much slower than YOLOv3.
  
YOLOv3 solves the detection task as a regression problem by resizing the input image to a default size (416$\times$416 in our case) and dividing it in a grid for each output scale, where each grid cell yields in output an array whose shape is B$\times$(5+C), where B is the number of rectangular bounding boxes a cell can predict, 5 is for the number of bounding box attributes and the object confidence, and C is the number of classes. Non Maximal Suppression (NMS) is finally used to keep only the predicted bounding boxes having highest confidence.

\subsection{Edges detection and image rotation based on Computer Vision}
YOLOv3's localization performances are optimal if target edges are parallel to the image edges in order to maximize the intersection between ground truths and detected bounding boxes.
We apply therefore a classical edge detection method to find the panel edges, get the predominant directions and rotate the image accordingly.
In particular, the procedure is composed by three main steps: (i) edges detection based on Canny detector, (ii) lines detection by means of Hough Transform, (iii) image rotation.

\subsubsection{Edges detection}
We adopted the well-known Canny edge detector \cite{Canny 86} which represents a good compromise
between accuracy, computational time and algorithm complexity. 
It is a multi-stages algorithm composed by the following steps:
\begin{enumerate}[{(i)}]
\item \textit{Noise reduction}: a 5x5 Gaussian filter is exploited to smooth the input image
\item \textit{Computation of the image intensity gradient}: a Sobel operator is applied to the smoothened image to get the first derivative of the image intensity in both horizontal and vertical directions, i.e. $G_x$ and $G_y$, respectively. From the two resulting images, edge gradient magnitude $G$ and direction $\alpha$ for each pixel are computed as follows:
\begin{equation}\label{edge_gradient}
G= \sqrt{G_x^2 + G_y^2} \;\;\; ; \;\;\; \alpha= tan^{-1}(G_x^2 / G_y^2)
\end{equation}
\item \textit{Non Maximum Suppression}: candidate edges correspond to points where the gradient is maximum.
\item \textit{Hysteresis thresholding}: a double-threshold image binarization method is applied with thresholds $ G_{min} < G_{max}$.
All the candidate edges having $G > G_{max}$ ($G < G_{min}$) are retained (discarded). The remaining edges are maintained only if connected to edges having $G > G_{max}$.  
\end{enumerate}
\begin{figure}[!tb]
\centering
	\vspace{0cm}
	\includegraphics[width=\linewidth, keepaspectratio]{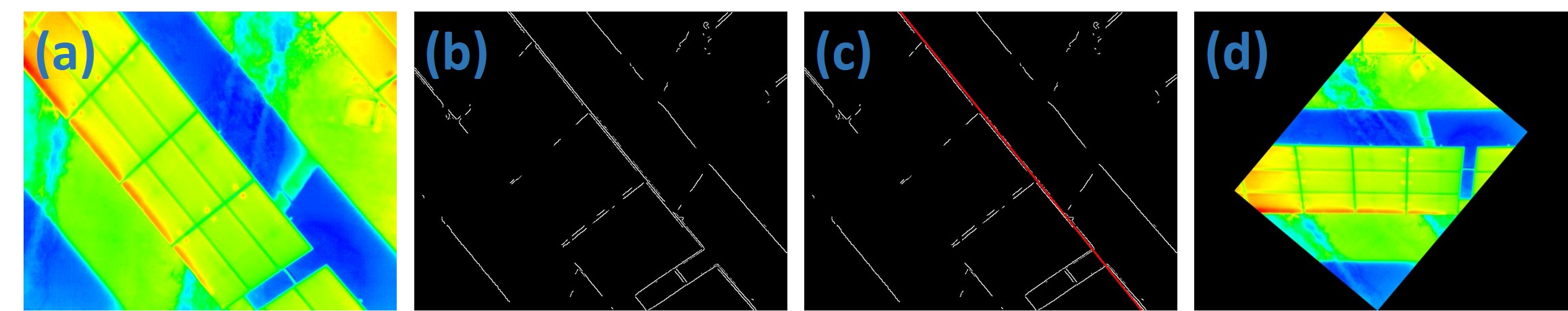}
	\vspace{0cm}	
	\caption{Image rotation methodology: (a) input image, (b) edge detection based on Canny detector, (c) predominant linear edge (red) identified by Hough transform, (d) rotated image.}
	\label{fig:cv_procedure}
\end{figure}
Starting from input image shown in Fig.~\ref{fig:cv_procedure}a, we obtain the binary image depicted in Fig.~\ref{fig:cv_procedure}b, where the white pixels set to 1 correspond to detected edges.

The two thresholds above $G_{min}$ and $G_{max}$, which mostly influence the number of detected edges, were set empirically by a trial and error approach on training set to default values of 450 and 550, respectively.
Moreover, to face the issue of missing edges in some circumstances, we designed an iterative in-house algorithm that dynamically changed the thresholds depending on the input image. In brief, starting from the values above, the number of detected edges $N_{edges}$ are counted and, if $ N_{edges} < N_{thr}$, then $G_{min}$ and $G_{max}$ are decreased by 50, and the edge detection is repeated. 
The procedure is stopped once a suitable number of pixels set to 1 is achieved, i.e. $ N_{edges} >= N_{thr}$.
Finally, to set the optimal value of $N_{thr}$, we apply the inflection point (Elbow) method on the distribution of sorted increasing values of $N_{edges}$ computed on training images by using default values for $G_{min}$ and $G_{max}$. We found $N_{thr}= 1000$.  


\subsubsection{Lines detection}
In order to detect linear edges, we applied the Hough Transform \cite{Shehata 15} on the binary image 
presented in Fig.~\ref{fig:cv_procedure}b. 
Hough transform maps the edge points into cosine curves in the Hough Space $(\rho, \theta)$ and detected lines correspond to $(\rho, \theta)$ pairs having a number of intersections larger than a threshold. 
In order to detect only one predominant direction suitable to rotate the image, we implemented an iterative approach starting from a high threshold and decreasing progressively its value until one line and the corresponding angle with respect to the horizontal axis are returned.
An example of the result achieved is shown in Fig.~\ref{fig:cv_procedure}c.

\subsubsection{Image rotation}
Finally, we applied a geometrical rotation to the input image based on the computed angle.
Fig.~\ref{fig:cv_procedure} summarizes from left to right the steps followed to rotate the input image.

\subsection{Defect Detector}
YOLOv3 requires a list of images and the corresponding set of ground truth boxes and class labels to learn. Annotation has been realized by means of the open LabelImg graphical user interface~\cite{LabelImg}: it has been done at instance level in IR images and either at instance or panel level, depending on class, in VIS images (section \ref{Datasets}).  
Consequently, the defect detector includes a preliminary VIS image rotation to better fit anchors on ground truths during YOLOv3's training and inference (Fig.~\ref{fig:model_pipeline}). 

Due to the complexity of YOLOv3, which includes more than 60 millions of parameters, training from scratch is usually avoided since the amount of data necessary is prohibitive leading to the overfitting problem caused by data insufficiency. 
Hence, transfer learning technique starting from pre-trained weights on the publicly available COCO dataset~\cite{Lin 14} has been preferred. This allows to effectively transfer the knowledge from the original detection task, i.e. detection on COCO, to the new PV domain, while handling with a limited amount of data. It also speeds up learning. Additionally, this is particularly convenient in our case where some defect classes exhibit a few hundred instances (Tables~\ref{tab:statistics_sicilia_dataset} and~\ref{tab:statistics_campania_dataset_class}).

To further increase the statistics available during fine-tuning, Data Augmentation (DA) based on geometrical transformations have been applied to training data (Tab.~\ref{tab:data_augmentation_defect}). Optical distortion acting in the HSV spectrum was instead not modelled as it could weaken the distinctive color based features of some defect classes.
Moreover, since a preliminary rotation was applied to VIS images before training, rotation DA was applied only to IR images.

During inference, detections are returned in the VOC format in a csv file, i.e. for each input image an output file is produced containing in each row the class label, the confidence score of the prediction, as well as the top left and the bottom right corners of the bounding box expressed in pixel units.

\begin{table}[h]
\small
\centering
\begin{tabular}{c|c|c}
\hline                       
\rowcolor[gray]{0.9} \textbf{Data Augmentation} & \textbf{IR images} & \textbf{VIS images} \\
\hline                       
rotation & \checkmark &  \\
horizontal flip & \checkmark \xmark &  \checkmark \xmark \\
vertical flip & \checkmark \xmark &  \checkmark \xmark\\
cropping & \checkmark &  \checkmark\\
scaling & \checkmark &  \checkmark\\
\hline 
\end{tabular} 
\caption{\label{tab:data_augmentation_defect}Dataset Augmentation techniques implemented during training for IR images (Plant\textunderscore Sicilia) and VIS images (Plant\textunderscore Campania), and for defect (\checkmark) and panel (\xmark) detectors, respectively.}
\end{table}

\subsection{Panel detector} 
To improve the YOLOv3 detection task, the panel detector first applies a preliminary rotation aimed to orientate PV panels either vertically or horizontally (Fig.~\ref{fig:model_pipeline}) before learning or inference. DA techniques during training do not include therefore rotation, nor scaling since panel dimensions are roughly similar (Tab.~\ref{tab:data_augmentation_defect}).  

\begin{figure}[!tb]
\centering
	\vspace{0cm}
	\includegraphics[width= \linewidth, keepaspectratio]{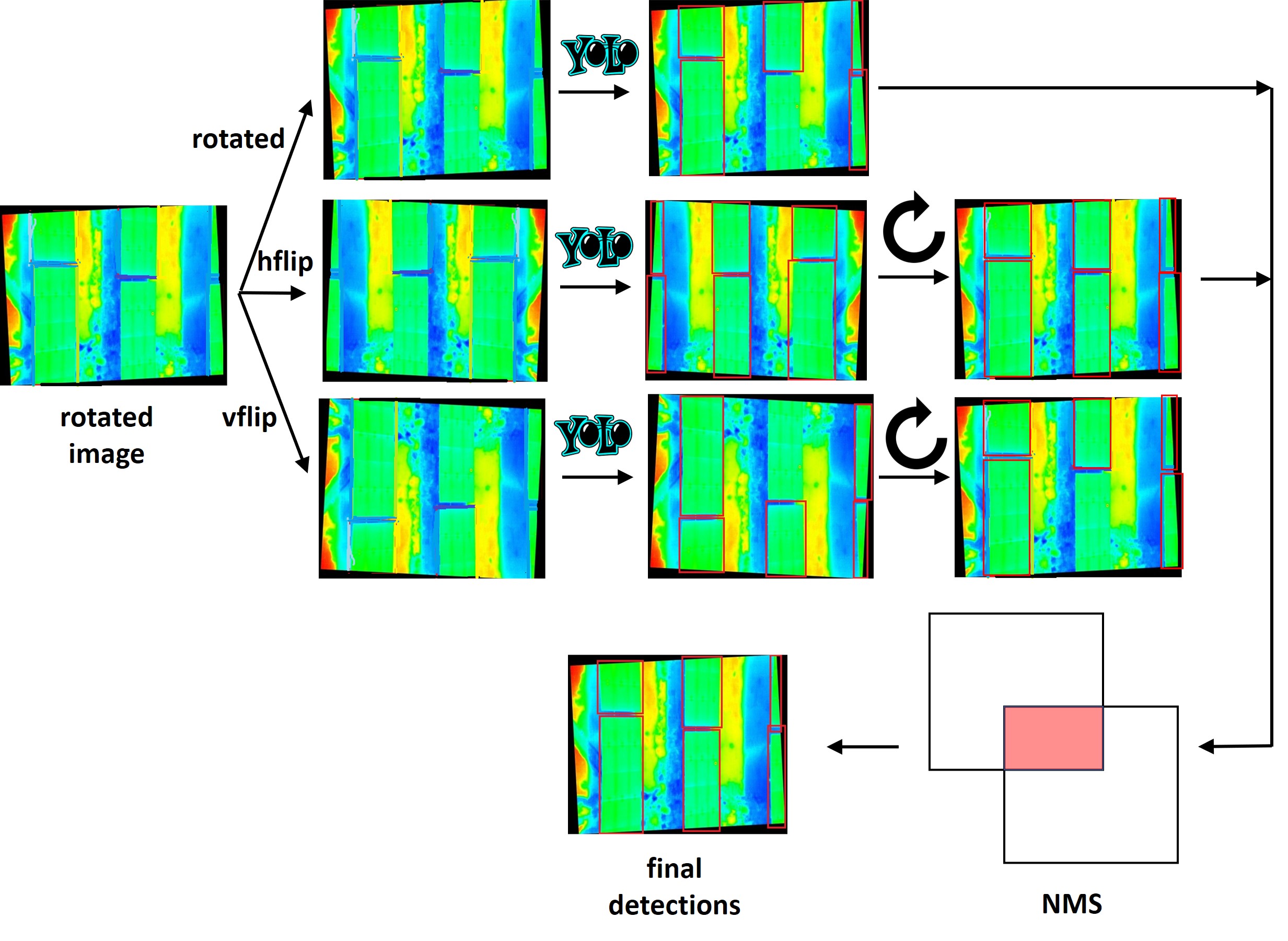}
	\vspace{0cm}	
	\caption{Test Time Augmentation (TTA) workflow: the initial rotated image is augmented by horizontal and vertical flip, then inference based on YOLOv3 is applied on the three images. Finally detections are expressed in the same reference system, pruned by means of NMS and returned.}
	\label{fig:tta_procedure}
\end{figure}

In addition, at inference time, to enhance YOLOv3 performances, test Time Augmentation (TTA), in a manner similar to Ref.~\cite{Garcia 20}, is applied to the rotated image generating two further images horizontally and vertically flipped.  
Then inference is executed on the three resulting images.
Finally the detections are pruned according to NMS algorithm and the detected PV panels are returned.
The TTA procedure is shown schematically in Fig.~\ref{fig:tta_procedure}, whereas the post-processing NMS proceeds as follows: we consider as real bounding boxes the set $BB_R= \left\{bb_1^R, ..., bb_k^R\right\}$ produced on the original rotated image and as candidate detections all the others $BB_C= \left\{bb_1^C, ..., bb_n^C\right\}$ related to the augmented images.
Then we sort $BB_C$ in decreasing order of box confidence score, we pick from $BB_C$ the box $bb_i^C$ having highest confidence and we compute its Intersection over Union (IoU)
\begin{equation}\label{iou}
IoU\left(bb_i^C, bb_j^R\right)= \frac{bb_i^C \cap bb_j^R}{bb_i^C \cup bb_j^R} 
\end{equation}
with each box $bb_j^R \in BB_R$ . If $IoU\left(bb_i^C, bb_j^R\right) < IoU_{thr}^P \; \forall  bb_j^R \in BB_R$, then $BB_R= BB_R \cup bb_i^C$ and $BB_C= BB_C \setminus bb_i^C$.
Otherwise we discard $bb_i^C$ and we remove it from $BB_C$.
NMS continues until $BB_C$ is not empty. In this work, $IoU_{thr}^P$ has been set empirically to 0.2 in order to allow a minimum overlapping between panel proposals.
The detections of panel detector are finally delivered in the same format as defect detector.

\subsection{False Alarm Filter} 
False Alarm Filter (Fig.~\ref{fig:model_pipeline}) is fed with the following: (i) defect candidates by defect detector and (ii) panel detections by the panel detector.
Once the defect candidates have been expressed in the same reference system of panel detections (this actually requires a rotation for IR images), then the IoU between the defect and panel bounding boxes is computed. If $IoU > IoU_{thr}^D$, then the defect is retained, otherwise is discarded. Since the defect bounding box may exceed the border of the panel proposal, we set empirically $IoU_{thr}^D= 0.5$.
Finally the false alarm filter returns the defect detection expressed in the reference system of the input image. 

In addition, in case of thermal images (Plant\_{Sicilia}) the hotspot severity and the corresponding prescribed action for O\&M operators are delivered.
On the other hand, in case of VIS imaging (Plant\_{Campania}), if strong soiling is detected, the soiling coverage is also returned by computing the intersection area between the defect bounding box $bb_{def}$ and the panel box $bb_{pan}$ where the defect is located into, i.e. $\left(bb_{def} \cap bb_{pan}\right)/bb_{pan}$. Here below an example of delivered textual file for each input VIS image is shown:\\
\textit{
dataset/test\textunderscore set/15-03-05-294\textunderscore digital.jpg\\
Panel 5: strong soiling covers 27.22 \% of the whole area\\
Panel 6: strong soiling covers 7.81 \% of the whole area\\
Panel 12: strong soiling covers 10.68 \% of the whole area\\
}
More details about the automatic analysis of temperature gradient due to hotspots are provided in the next section.

\subsection{Prediction of hotspot severity based on thermal analysis for Plant\_Sicilia} 

We remark that in this work the power loss due to a defective module is not estimated, unlike Ref.~\cite{Mehta 18}, due to unavailability of electrical tags collected by on-site data-logger. An alternative approach based only on analysis of thermal images has been presented so far by Ref.~\cite{Denz 20}, based on the assumption of a steady state close system where all the energy that is not converted to electrical power is dissipated as heat through the Joule effect, resulting proportional to the temperature difference $\Delta T$ between the defective (D) module and a neighbour normal (N) operating module, i.e. $P_{loss}= k \cdot \Delta T= k \cdot (T_D - T_N)$.

In this case we have not followed such method for two main reasons. First, the coefficient $k$ requires an on-site calibration which herein was not available: indeed in~\cite{Denz 20} it has been estimated by the knowledge of a short-circuit module dissipating all the electrical power into heat. However, depending on the magnitude of the estimated slope $k$, the predicted power loss may vary significantly. 
Second, when estimating $P_{loss}$, temperature-only analysis is often not sufficient since multiple factors can affect power loss as the type of the PV module, its efficiency, the mounting system, its orientation, the geographical position and, above all, the electric interconnection and conversion plant. Actually, O\&M asset managers are often more interested to know how many modules must be replaced based on analysis of temperature gradient between hotspot area and healthy surrounding nominal cells.
Indeed this information allows O\&M operators to identify real short-circuits cases comparing it with predefined thresholds based on IEA's recommendations~\cite{IEA 18}, as for example discussed in Ref.~\cite{Saavedra 19}.

As a consequence, in this work the model does not predict the power loss, but it automatically estimates the temperature gradient induced by the hotspot and the related severity, like in Ref.~\cite{Saavedra 19}. 
In particular, the model automatically extracts the maximum hotspot temperature from the detected area and estimate the average temperature of a neighbour healthy PV module in the same PV array based on the detected panel area and edge detection method. Then it automatically leverages the temperature gradient to estimate the defect severity, and prescripts the corresponding suggested intervention to O\&M operators aimed to minimize cost of PV plant maintenance, ensure safe operation and maximize yield, according to recommendations provided by Ref.~\cite{IEA 18}. 

More in detail, the computed temperature gradients $\Delta T$ are grouped in four categories of increasing severity (Tab.~\ref{tab:temperature_analysis}), in a way similar to Ref.~\cite{Saavedra 19}: $\Delta T$ smaller 10~\textcelsius~ is considered normal behaviour, values between 10~\textcelsius~ and 20~\textcelsius~ correspond to heated cells which must be paid attention to in regular thermographic inspections, whereas $\Delta T$ above 20~\textcelsius~ refers to critical hotspots which can cause severe degradation in module generation, as well as safety issue during maintenance work. Hence PV module replacement is recommended, as also suggested in Ref.~\cite{Libra 19}. In this work we also discriminate between gradients in the interval [20, 30)~\textcelsius~ (severe) and above 30~\textcelsius~ (extremely severe) in order to better prioritize maintenance interventions.

\begin{table}[h]
\small
\centering
\begin{tabular}{c|c|c|c}
\hline                       
\rowcolor[gray]{0.9} \textbf{Defect} & \textbf{Temperature} & \textbf{Recommended action} \\
\rowcolor[gray]{0.9} \textbf{severity} & \textbf{difference [\textcelsius]} &  \\
\hline                       
Normal & $0\leq \Delta T < 10$ & None \\
\hline                       
Heated Cell(s) & $10\leq \Delta T < 20$ & Careful check in regular thermographic inspections \\
\hline                       
Severe hotspot & $20\leq \Delta T < 30$ & Replacement of the defective module \\
\hline                       
Extremely Severe & $ \Delta T \geq 30$ & Immediate replacement of the defective module \\
hotspot &  &   \\
\hline 
\end{tabular} 
\caption{\label{tab:temperature_analysis}Classification of defect severity based on temperature gradient $\Delta T$ induced by the defect and related action to solve the issue suggested to O\&M operator based on~\cite{IEA 18}.}
\end{table}

\subsection{k-Means Clustering} 
YOLOv2 introduces the concept of anchors~\cite{Redmon 16b}, a set of initial box candidates which are adjusted by the network during learning. Box priors may be fitted on training dataset to improve accuracy and speed. Unsupervised k-means clustering can be used to automate the procedure. However, clustering based on classical Euclidean distance weights larger boxes more then smaller ones for comparable values of overlapping.
Therefore we adopt an IoU-based distance function:
\begin{equation}\label{kmeans_metric}
d\left(a,b\right)= 1-IoU\left(a,b\right) 
\end{equation} 
where $a$ is the ground truth box and $b$ is the anchor box. To determine good priors, we applied k-means on training set and we monitored the average IoU, as well as the mean Silhouette and the total intra-cluster variation as a function of the number of clusters $k$.
The total intra-cluster variation measures the compactness of the clustering and it is computed as the sum of squared distance errors (SSE) between a data point $x_i \in C_j$ and the centroid $\mu_j$ of cluster $C_j$~\cite{Nainggolan 19}, i.e.:
\begin{equation}\label{sse}
SSE= \sum_{j=1}^k \sum_{x_i \in C_j} ||x_i - \mu_j||^2 \; .
\end{equation} 
The silhouette coefficient $S$ represents instead the degree of inter-cluster separation and is evaluated for each instance as~\cite{Sudheera 16}:
\begin{equation}\label{silhouette}
S_i= \frac{b_i - a_i}{max\left(a_i, b_i\right)} \; ,
\end{equation} 
where $a_i$ ($b_i$) is the average intra-distance (inter-distance) of the $i$-th sample from all points in the same (closest) cluster ($ -1 \leqslant S_i \leqslant 1$). The better the clustering, the larger the value of the mean silhouette $\sum_{i=1}^N S_i / N$ and the smaller the SSE over $N$ samples of training dataset. 

\begin{figure}[!tb]
\centering
	\vspace{0cm}
	\includegraphics[width= \linewidth, keepaspectratio]{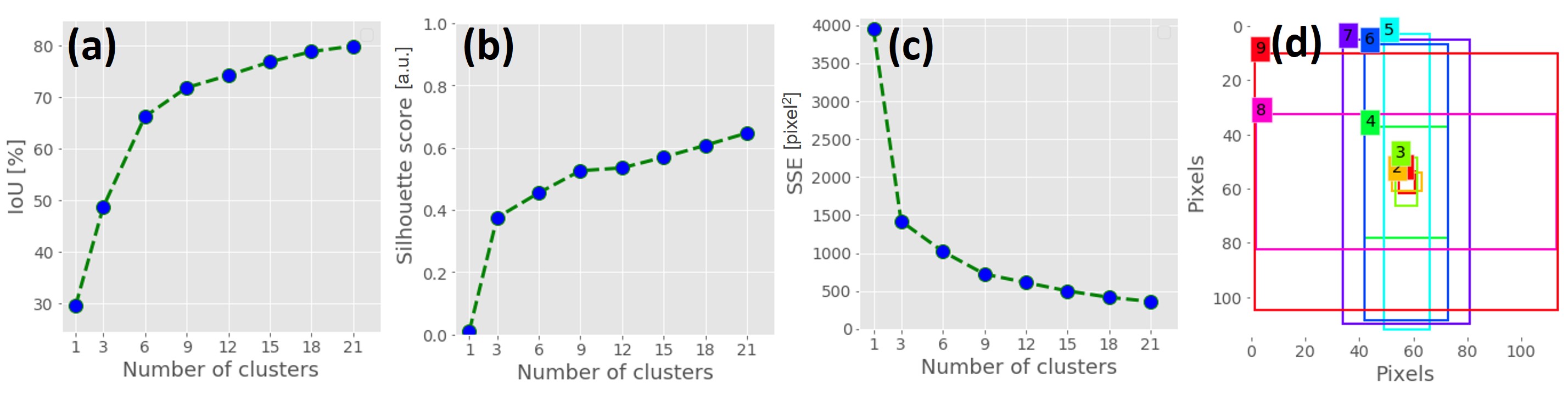}
	\vspace{0cm}	
	\caption{k-means clustering applied to training dataset of defect detector for Plant\_Campania: (a) IoU, (b) silhouette and (c) SSE as a function of number of clusters. (d): resulting anchors.}
	\label{fig:kmeans_campania_defect}
\end{figure}

Fig.~\ref{fig:kmeans_campania_defect}a-c shows these quantities for the defect detector of Plant\textunderscore Campania: according to the Elbow method, we selected $k=9$ clusters as the best trade-off between accuracy and speed. Then we divided up the clusters on three scales, i.e. $B= k/3$. Fig.~\ref{fig:kmeans_campania_defect}d shows the resulting anchor boxes, corresponding to the centroids of the clusters. The procedure has been repeated similarly in the other cases.

\section{Results} \label{Results}
\subsection{Metrics}
In this work we used the PASCAL VOC metrics~\cite{Everingham 15} to evaluate performances.
Given an IoU threshold, we compute Recall as the proportion of correctly detected boxes $REC= \left(TP\right)/\left(TP+FN\right)$, Precision as the ratio between the correctly detected boxes and the overall number of detected boxes $PREC= \left(TP\right)/\left(TP+FP\right)$ and F1-score as the harmonic mean $F1= 2*\left(PREC \cdot REC\right)/\left(PREC + REC\right)$. Here TP, FP and FN denote the number of True Positives, False Positives and False Negatives, respectively. Then for each class we draw the Precision-Recall curve and we evaluate the Average Precision (AP) as the area under the curve:
\begin{equation}\label{ap}
AP = \int^1_0 PREC(REC)\,dREC
\end{equation}
Finally the mean Average Precision is computed as the arithmetic mean of the AP values of the different classes
\begin{equation}\label{map}
mAP = \frac{\sum_{i=1}^{n}AP_i}{n}
\end{equation}
where $n$ is the number of classes under consideration.

\subsection{Training}
\begin{figure}[!tb]
\centering
	\vspace{0cm}
	\includegraphics[width= \linewidth, keepaspectratio]{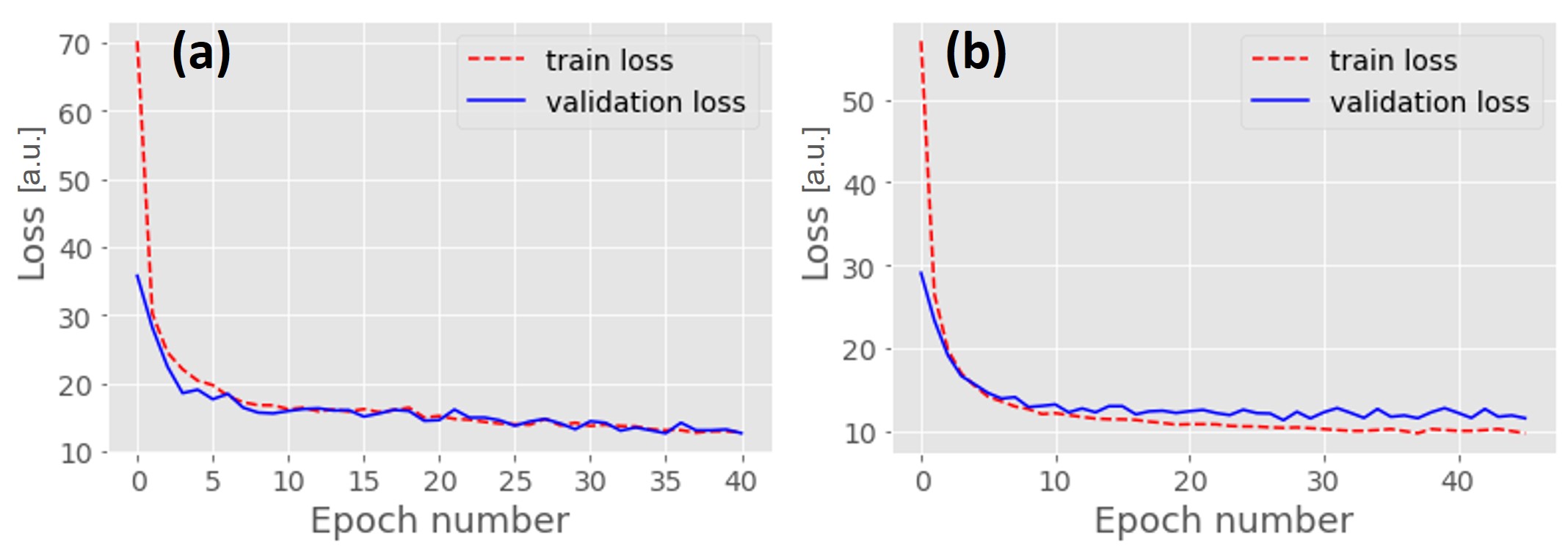}
	\vspace{0cm}	
	\caption{Training (red) and validation (blue) losses for (a) defect detector and (b) panel detector. The PV system under consideration is Plant\_Sicilia.}
	\label{fig:loss_sicilia}
\end{figure}

The proposed model has been implemented by means of the Keras library within the TensorFlow framework. The experiments were conducted using an Intel Xeon E5-2690 v3 processor with 56 GB RAM and NVIDIA Tesla K80 GPU.
To handle the problem of data lack, as well as to speed up learning and improve accuracy, a 2-stage training starting from pre-trained weights on COCO dataset~\cite{Lin 14} has been adopted: initially transfer learning has been executed for 3 epochs by unfreezing the last three layers of the network and using a learning rate of 0.001 and a batch size of 32. Then the network has been finetuned by training the whole architecture and using a learning rate of 0.0001 and a batch size of 8. 
To avoid overfitting, we implemented early-stopping by monitoring validation loss and stopping finetuning once no more loss decrease was observed for ten consecutive epochs.
As revealed in Fig.~\ref{fig:loss_sicilia} for Plant\_Sicilia, the convergence of training and validation loss functions for defect and panel detectors is fast and reached after few tens of epochs.

In addition, monitoring of learning curves denotes also that the learned model does not neither overfit, nor underfit and confirm the representativeness of the built training and validation sets. In fact, the learning curves continue to decrease smoothly until convergence is reached, with a negligible gap between the two curves, proving that the detectors are good fits. On the other hand, the models generalize well on the unknown validation set and validation loss converges close to the training loss, demonstrating both that training set is representative of the problem under investigation, as well as that the validation set is a good benchmark for performance evaluation, respectively.

\subsection{Results for Plant\_Sicilia}
\subsubsection{Panel Detector}
Tab.~\ref{tab:tta_sicilia} shows the margin of improvement achieved for Plant\_Sicilia @0.5 when applying TTA during inference. TTA increases obviously the overall number of correct detections and misdetections. The overall effect is an AP improvement of roughly 0.64\%. 
Tab.~\ref{tab:performance_panel_sicilia} shows the evaluation metrics for different IoU on test set and results from Ref.~\cite{Herraiz 20}: as can be seen, our panel detector outperforms Ref.~\cite{Herraiz 20} for all the considered localization thresholds. In particular, our REC@0.5 is more than 10\% higher than in Ref.~\cite{Herraiz 20}: this is a remarkable result since a high false negative rate may negatively influence the hotspot detector outcome as hotspots identified outside detected panel areas are removed in both models. Globally, our AP@0.5 is almost 98.5\% and well above 90\% also for higher IoU. Only 13 out of 926 panels are missed @0.5, usually not appearing completely in the image, as also observed in Ref.~\cite{Herraiz 20}. 

\begin{figure}[!tb]
\centering
	\vspace{0cm}
	\includegraphics[width= \linewidth, keepaspectratio]{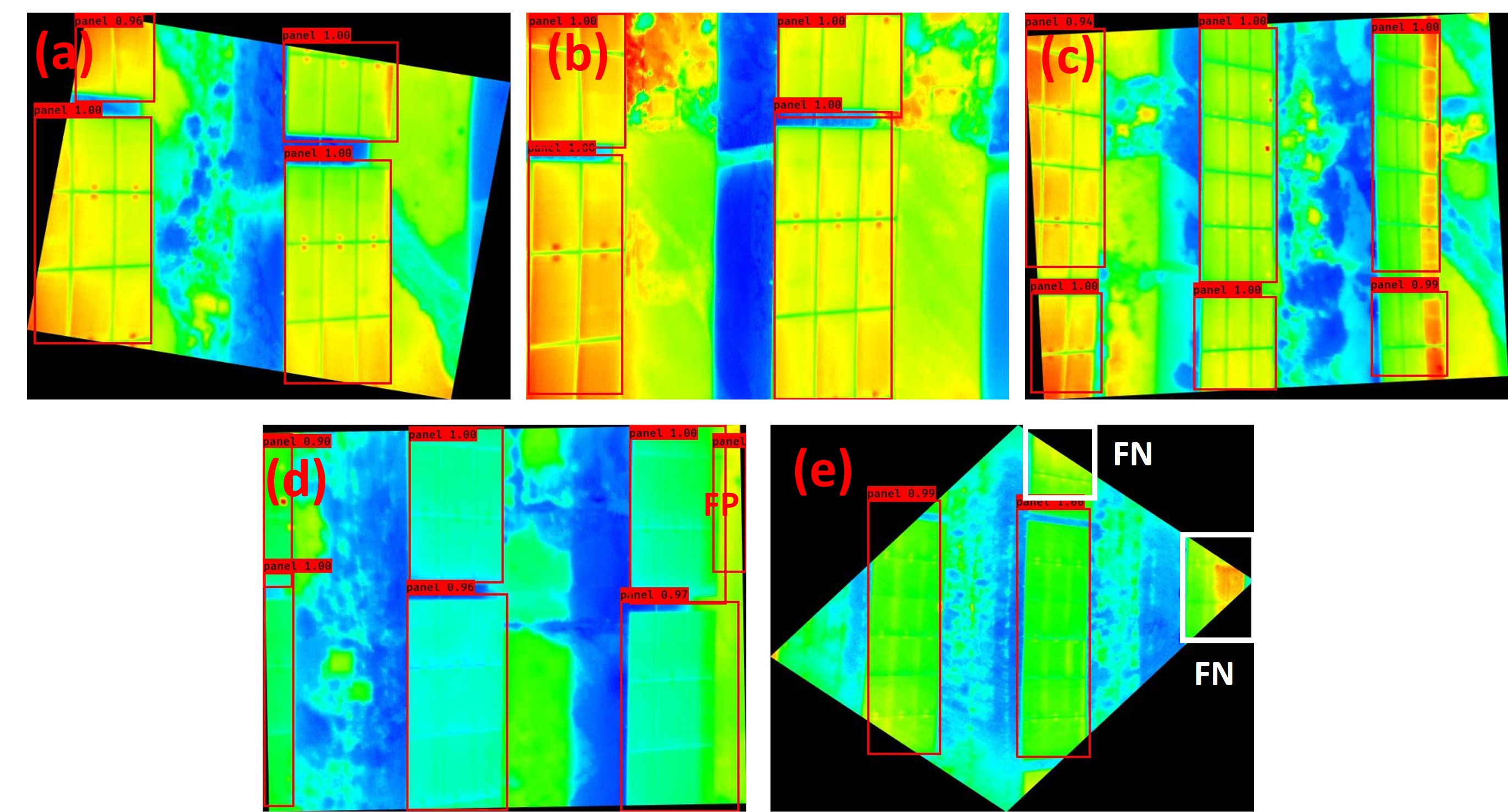}
	\caption{Examples of detections (red boxes) made by the panel detector on test images of Plant\_Sicilia. (a), (b), (c): TPs; (d): FP; (e): FN (white box).}
	\label{fig:example_panel_sicilia}
\end{figure}

Fig.~\ref{fig:example_panel_sicilia} shows some detection examples. As may be seen in Fig.~\ref{fig:example_panel_sicilia}a-c, the model generalizes well for different panel orientation and size, ensuring also a classification with a very high level of confidence (usually not less than 0.95).
FPs are very rare and usually correspond to the detection of a small part of a panel at the image border (Fig.~\ref{fig:example_panel_sicilia}d), thus not penalizing the outcome of the false alarm filter. FNs are equally rare and related to panel portions located at the boundary and with a significant slope with respect to the horizontal in the input image (Fig.~\ref{fig:example_panel_sicilia}e).


\begin{table}[h]
\small
\centering
\begin{tabular}{c|c|c}
\hline
\rowcolor[gray]{0.9} \textbf{Metric} & \textbf{without TTA} & \textbf{with TTA}  \\
\hline
PREC [\%] & \textbf{98.44} & 97.49 \\
REC [\%] & 98.33 & \textbf{99.00} \\
AP [\%] & 98.27 & \textbf{98.91}\\
\hline
\end{tabular}
\caption{\label{tab:tta_sicilia} Precision, Recall and AP @0.5 of panel detector on validation set of Plant\_Sicilia with and without the application of TTA during inference.}
\end{table}

\begin{table}[h]
\small
\centering
\begin{tabular}{c|c|c|c|c|c|c|c}
\hline
\rowcolor[gray]{0.9} \textbf{IoU} & \textbf{TP} & \textbf{FP} & \textbf{FN} & \textbf{PREC} [\%] & \textbf{REC} [\%] & \textbf{F1} [\%] & \textbf{AP} [\%] \\
\hline
0.3 & 915 & 11 & 11 & 98.81 & 98.81 & 98.81 & 98.71 \\
0.5 & 913 & 13 & 13 & \textbf{98.59} & \textbf{98.59} & \textbf{98.59} & \textbf{98.48} \\
0.7 & 877 & 49 & 49 & 94.70 & 94.70 & 94.70 &  93.97 \\
\hline
\rowcolor[rgb]{0.88,1,1} cascade~\cite{Herraiz 20} & NA & NA & NA & 94.52 & 88.40 & 91.35 & NA \\
\hline
\end{tabular}
\caption{\label{tab:performance_panel_sicilia} Evaluation metrics and absolute number of TP, FP and FN of panel detector for different IoU on test set of Plant\_Sicilia. Results from~\cite{Herraiz 20} are also reported.}
\end{table}

\subsubsection{Defect detector and multi-stage model}
\begin{figure}[!tb]
\centering
	\vspace{0cm}
	\includegraphics[width= \linewidth, keepaspectratio]{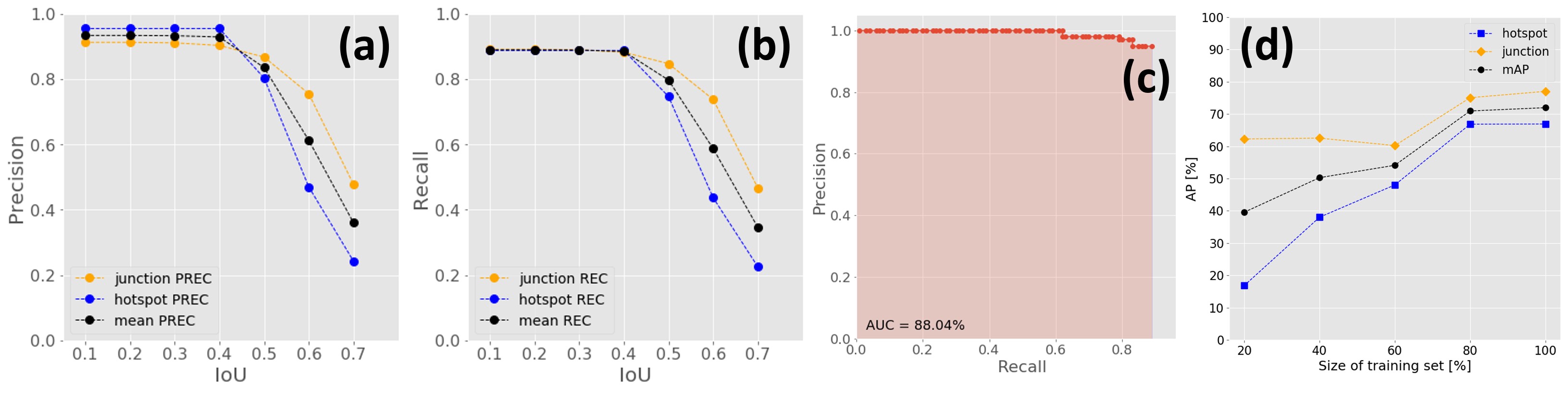}
	\caption{(a) Precision and (b) Recall as a function of IoU for hotspot and junction classes; (c) Precision-Recall curve for hotspot class @0.4; (d) APs and mAP @0.5 as a function of the training set size, expressed as a percentage of the full data amount available. Results correspond to test set of defect detector for Plant\_Sicilia.}
	\label{fig:hotspot_sicilia_defect_detector}
\end{figure}

Fig.~\ref{fig:hotspot_sicilia_defect_detector} depicts the relationship between recall and precision for classes hotspot and junction on test set of defect detector.
As can be seen, IoU= 0.4 is a good balance point for hotspot class, ensuring a REC (PREC) as high as 88.7\% (95.5\%), corresponding to only 8 FNs out of 71 hotspot instances and 3 FPs. Performances degrade progressively for higher localization thresholds and, as expected, the effect is more pronounced for hotspot than junction class due to their tiny size which may lead to weak features extracted after many convolution operations in the backbone Darknet-53.
Numerical results @0.4 are reported in Tab.~\ref{tab:performance_defect_sicilia} and, for a fair comparison with Literature, we also show metrics @0.5. 

A heuristic approach has been also followed to investigate the reliability and representativeness of the collected training set to generalize properly on a test sample drawn from the same scenario. Specifically, data has been partitioned into $N$ subsets S$_1$, S$_2$, …, S$_N$ with a strict inclusion relationship, i.e. $S_1 \subset S_2 \subset … \subset S_N$,  where S$_i$ ($i= 1, ..., N$) corresponds to $100 \cdot i/N$ \% of the overall training set (S$_N$) used in the experiments. 
If the adopted training set was adequately representative to generalize efficiently on unknown test samples, we would observe a progressively decrease of model accuracy sensitiveness to training data amount with a value of $i$ approaching towards $N$. 
In Fig.~\ref{fig:hotspot_sicilia_defect_detector}d the case of $N= 5$ for Plant\_Sicilia is presented. As can be seen, mAP@0.5 is almost constant until the size of training set is not less than 80\% of the full size available, whereas for smaller statistics it degrades gradually, the model overfits on training data and the degradation is more pronounced for the hotspot class. 

\begin{table}[h]
\small
\centering
\begin{tabular}{c|c|c|c|c|c|c|c|c|c}
\hline
\rowcolor[gray]{0.9} \textbf{class} & \textbf{IoU} & \textbf{TP} & \textbf{FP} & \textbf{FN} & \textbf{PREC} & \textbf{REC} & \textbf{F1}& \textbf{AP} &  \textbf{mAP} \\
\rowcolor[gray]{0.9}  & & & & & [\%] & [\%] & [\%] & [\%] &  [\%] \\
\hline
hotspot & 0.4 & 63  & 3  & 8  & 95.45  & 88.73 & 91.97 & 88.04 & \multirow{2}{*}{85.48} \\
junction & 0.4 & 1142 & 123 & 153 & 90.27 & 87.18 & 89.21 & 82.93 &  \\
\hline
hotspot & 0.5 & 53  & 13  & 18  & 80.30  & 74.64 & 77.37 & 66.68 & \multirow{2}{*}{71.86} \\
junction & 0.5 & 1097 & 168 & 198 & 86.72 & 84.71 & 85.70 & 77.03 &  \\
\hline
\end{tabular}
\caption{\label{tab:performance_defect_sicilia} Evaluation metrics @0.4 and @0.5 on test set of defect detector for Plant\_Sicilia. The absolute number of TP, FP and FN is also reported.}
\end{table}

\begin{figure}[!tb]
\centering
	\vspace{0cm}
	\includegraphics[width= \linewidth, keepaspectratio]{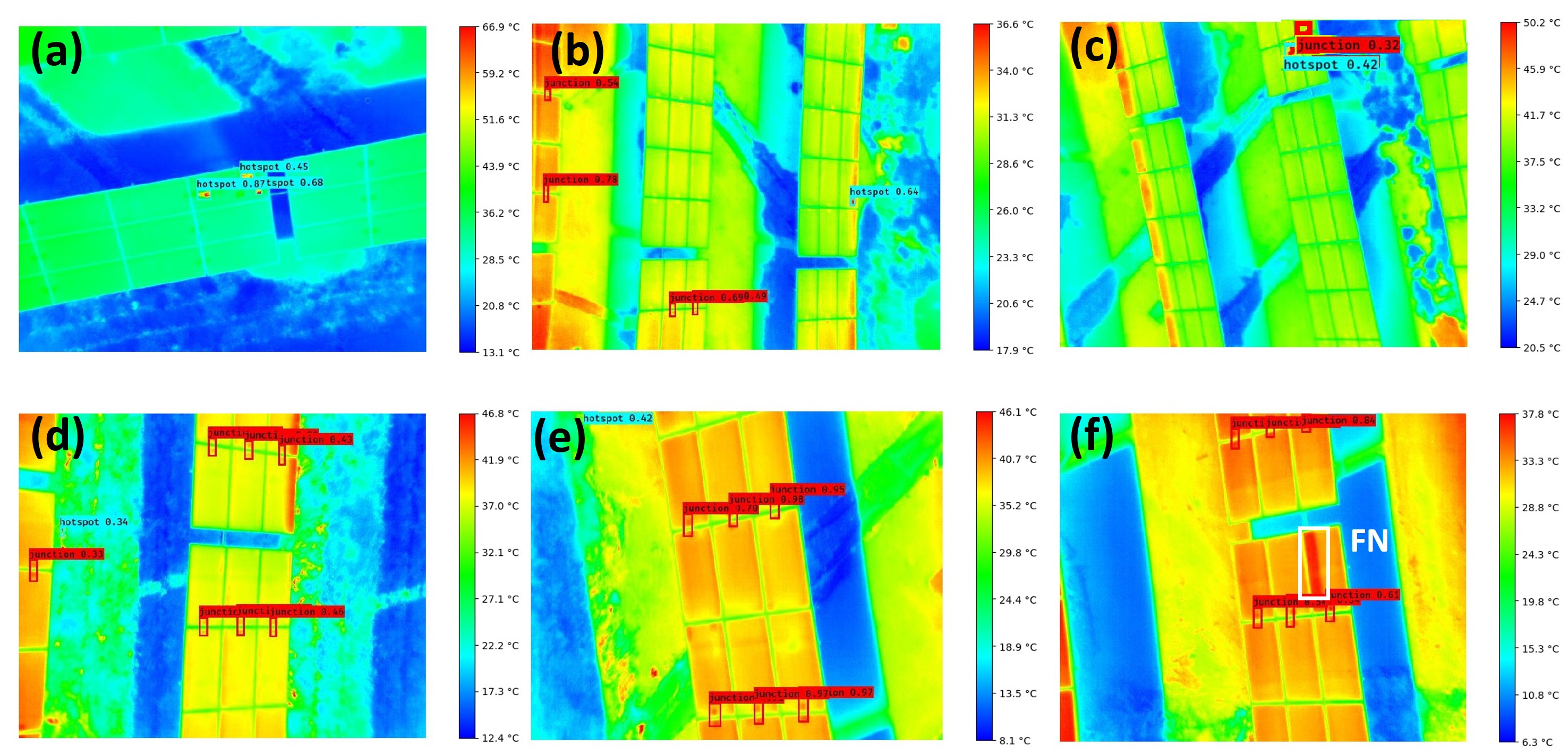}
	\caption{Examples of hotspot detections (sky blue boxes) made by the defect detector on test images of Plant\_Sicilia. (a), (b), (c): TPs; (d), (e): FPs, (f): FN (white box). In the same images the detections for junction box class are also superimposed (red boxes). The colorbar denotes the scene temperature expressed in units of Celsius degree.}
	\label{fig:example_hotspot_sicilia}
\end{figure}

Fig.~\ref{fig:example_hotspot_sicilia} shows instead some examples of detections. 
The defect detector is sensitive to small targets, as well as is able to discriminate accurately between hotspots and overheating due to junction box, in presence of cluttered background and changeable panel temperature (Fig.~\ref{fig:example_hotspot_sicilia}a-c).
The detector successfully identifies both hotspots much warmer than neighbourhood normal operating modules, with a temperature difference from about 10~\textcelsius~(Fig.~\ref{fig:example_hotspot_sicilia}c) to more than 30~\textcelsius~(Fig.~\ref{fig:example_hotspot_sicilia}a), and also hotspots only a few degrees hotter than nominal modules, with comparable temperature to junction heating (Fig.~\ref{fig:example_hotspot_sicilia}b).
The three misdetections @0.4 are either due to detections outside the panel area (Fig.~\ref{fig:example_hotspot_sicilia}d-e), thus removable by the false alarm filter, or to wide hot area at the panel edge.
FNs are instead caused either to unexpected hotspots with rectangular shapes (Fig.~\ref{fig:example_hotspot_sicilia}f) or with a smoother orange shade, both being not adequately represented in the training statistics.  
 
\begin{figure}[!tb]
\centering
	\vspace{0cm}
	\includegraphics[width= 13cm, keepaspectratio]{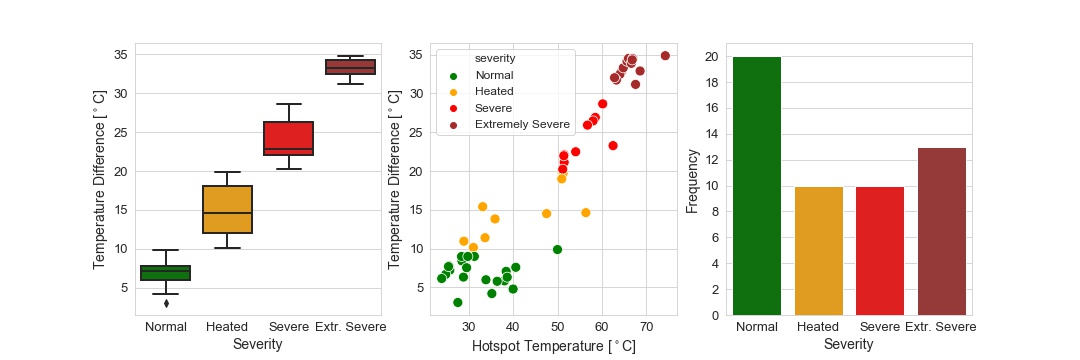}
	\caption{(a) Box chart of the temperature difference between hotspot and neighbour  healthy PV module grouped by defect severity; (b) scatter plot of temperature difference between hotspot and healthy area as a function of hotspot temperature; (c) frequency of detected hotspots grouped by severity. Data refer to TPs @0.5 of Plant\_Sicilia.}
	\label{fig:hotspot_analysis}
\end{figure} 

The model outputs also the temperature gradient $\Delta T$ between the detected hotspot and a neighbour healthy PV module of the same detected panel bounding box area, as well as the related severity and the suggested action to solve the issue based on the lookup Tab.~\ref{tab:temperature_analysis}. 
The model identifies 23 critical hotspots, requiring module substitution, out of 53 correctly identified hot areas @0.5 (Fig.~\ref{fig:hotspot_analysis}c). Of these, 13 defects have a $\Delta T$ over 30~\textcelsius, which corresponds to an estimated degradation of the energy conversion efficiency of more than 15\%~\cite{Libra 21}, and require immediate intervention. Severe anomalies have tipically  temperature exceeding 50~\textcelsius~ and up to more than 70~\textcelsius~ (Fig.~\ref{fig:hotspot_analysis}b). In particular, the median temperature gradients are almost 33.3~\textcelsius~ and 22.9~\textcelsius~ for extremely severe and severe hotspots, respectively, whereas is about 7.1~\textcelsius~ for nominal cells.

Tab.~\ref{tab:performance_multistage_sicilia} benchmarks the defect detector, the overall multi-stage model and results from Ref.~\cite{Herraiz 20}, which presented a dataset containing a similar statistics for hotspots. First, it is worth noticing that @0.4 the false alarm filter removes 2 out of 3 FPs, increasing PREC (F1-score) by almost 3\% (1.4\%), and therefore improving the overall robustness of the proposed model. 
Second, @0.4 the proposed model outperforms  Ref.~\cite{Herraiz 20} with a better F1-score almost 3\% higher, and with a similar REC. On the other hand, since Ref.~\cite{Herraiz 20} did not specify the IoU, if we assume IoU= 0.5, our model is significantly outperformed by Ref.~\cite{Herraiz 20} with a F1-score (REC) almost 12\% (15\%) lower. 
Since our panel detector has better performances, such degradation should be therefore imputed to the YOLOv3 engine used by hotspot detector, rather than to the false positive pruning stage: indeed Ref.~\cite{Herraiz 20} exploited the double-stage Faster R-CNN, which is slower than YOLOv3, but should be also more accurate in detecting small objects. 
\begin{table}[h]
\small
\centering
\begin{tabular}{c|c|c|c|c|c|c|c|c}
\hline
\rowcolor[gray]{0.9} \textbf{method} & \textbf{IoU} & \textbf{TP} & \textbf{FP} & \textbf{FN} & \textbf{PREC} & \textbf{REC} & \textbf{F1}& \textbf{AP} \\
\rowcolor[gray]{0.9}  & & & & & [\%] & [\%] & [\%] & [\%] \\
\hline
hotspot & 0.4 & 63  & 3  & 8  & 95.45  & 88.73 & 91.97 & 88.04 \\
\rowcolor[rgb]{ .886,  .937,  .855} multi-stage & 0.4 & 63  & 1  & 8  & \textbf{98.43}  & 88.73 & \textbf{93.33} & \textbf{88.31} \\
\hline
hotspot & 0.5 & 53  & 13  & 18  & 80.30  & 74.64 & 77.37 & 66.68 \\
\rowcolor[rgb]{ .886,  .937,  .855} multi-stage & 0.5 & 53  & 11  & 18  & \textbf{82.81}  & 74.64 & \textbf{78.51} & \textbf{66.91} \\
\hline
\rowcolor[rgb]{0.88,1,1} cascade~\cite{Herraiz 20} & NA & 77 & 7 & 9 & 91.67 & 89.53 & 90.58 & NA \\
\hline
\end{tabular}
\caption{\label{tab:performance_multistage_sicilia} Evaluation metrics and absolute number of TP, FP and FN @0.4 and @0.5 on test set of Plant\_Sicilia for defect detector and multi-stage model. Results from~\cite{Herraiz 20} are also shown.}
\end{table}

Tab.~\ref{tab:inference_time_sicilia} reports the timings of the automatics steps: panel detection is the most computationally demanding phase due to TTA, followed by image rotation, preliminary used to align panel to image edges, and false positive filtering, which carried out an initial alignment of defects to panel bounding boxes before pruning. Overall, the model takes less than 1~s to process one full-sized image including one or more PV arrays each. While this inference time is certainly only suitable for offline processing, it appears roughly one order of magnitude smaller than that achievable by first segmenting or cropping the input aerial image into patches of singular PV modules and then detecting anomalies~\cite{Bommes 21}. 
The method is also much more convenient than a manual post-processing, hence reducing effectively cost of O\&M operations: for example in~\cite{Saavedra 19} it took almost 26 working days to analyse thermal images of 17142 modules installed in a 3~MW PV plant.

\begin{table}[h]
\small
\centering
\begin{tabular}{c|c|c|c|c|c|c|c|c|c}
\hline
\rowcolor[gray]{0.9} \textbf{Detection Stage} & \textbf{Time per image [s]} \\
\hline
Panel Rotation & 0.248 \\
\hline
Panel Detection & 0.354 \\
\hline
Defect Detection & 0.130 \\
\hline
False Alarm Filter & 0.244 \\
\hline
\rowcolor[rgb]{0.88,1,1} Overall Time  & 0.976 \\
\end{tabular}
\caption{\label{tab:inference_time_sicilia} Average inference time per image for Plant\_Sicilia.}
\end{table}


\subsection{Results for Plant\_Campania}
\subsubsection{Panel Detector}
Tab.~\ref{tab:performance_panel_campania} shows the performance metrics for different localization thresholds on test set. Panel detector has an outstanding AP@0.5 of almost 97.9\%, with only 42 FN out of 2687 panel instances, and an AP above 90\% up to IoU of 0.7.

\begin{figure}[!tb]
\centering
	\vspace{0cm}
	\includegraphics[width= \linewidth, keepaspectratio]{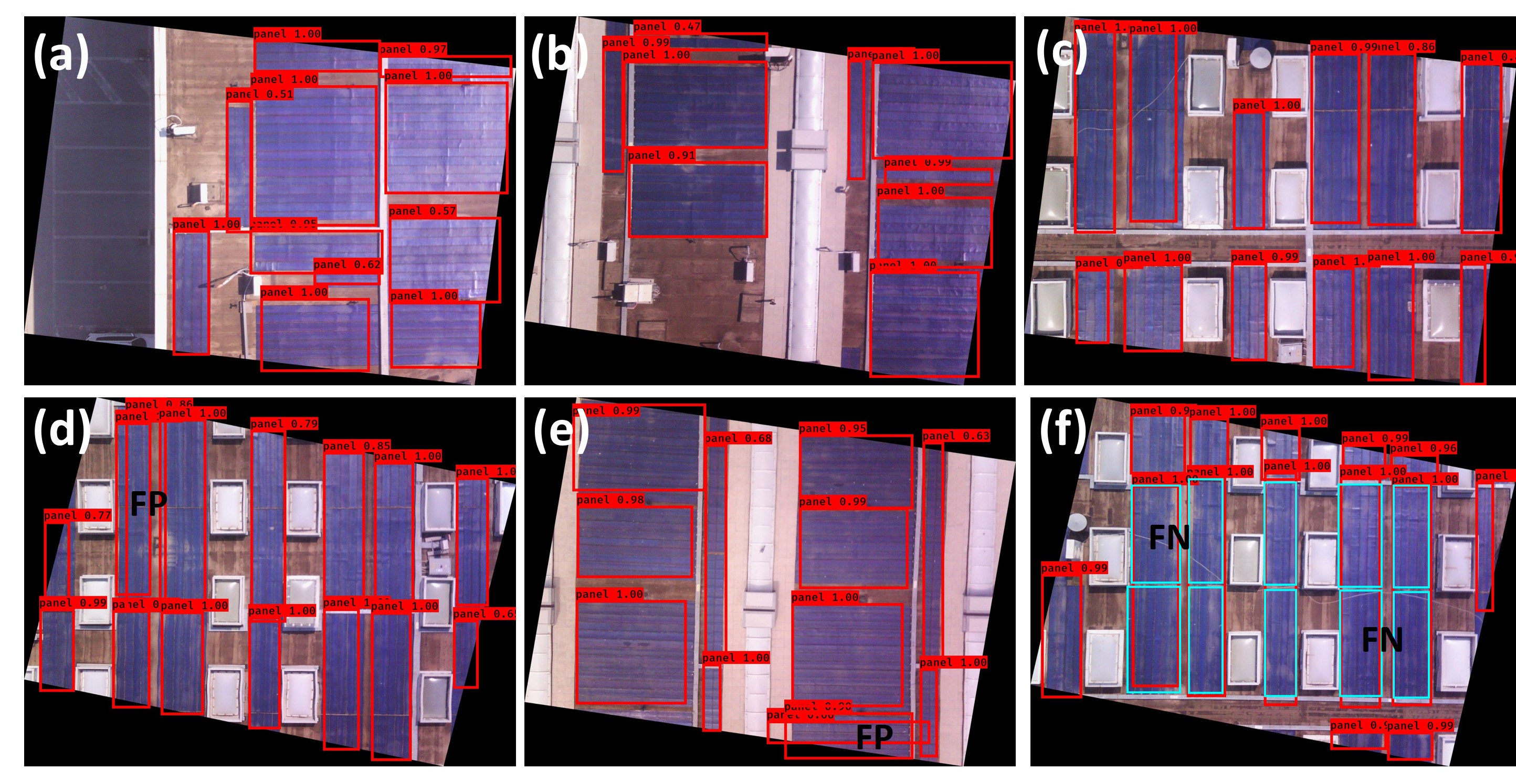}
	\caption{Examples of detections (red boxes) made by the panel detector on test images of Plant\_Campania. (a), (b), (c): TPs; (d), (e): FP; (f): FN and FP (in sky blue the GTBs).}
	\label{fig:example_panel_campania}
\end{figure}

As revealed by Fig.~\ref{fig:example_panel_campania}a-c, the model is sensitive to panels of different shape and orientation. Unlike Plant\_Sicilia, the confidence score may be as low as 0.45 due to the more significant variability in panel shape which makes detection harder.
Misdetections are more likely to appear at the image border and for small stripes, resulting in multiple detection for the same panel (Fig.~\ref{fig:example_panel_campania}d-e). Missed detections are rarer than FPs (Tab.~\ref{tab:performance_panel_campania}) and occur more often for elongated stripes which are detected as a whole (Fig.~\ref{fig:example_panel_campania}f), unlike annotation that segmented them into two parts due to a modest separation space. Actually both FPs and FNs depicted in  Fig.~\ref{fig:example_panel_campania} do not penalize the false alarm filter outcome as long as the panel area is detected and the distance between adjacent panels is negligible.

\begin{table}[h]
\small
\centering
\begin{tabular}{c|c|c|c|c|c|c|c}
\hline
\rowcolor[gray]{0.9} \textbf{IoU} & \textbf{TP} & \textbf{FP} & \textbf{FN} & \textbf{PREC} [\%] & \textbf{REC} [\%] & \textbf{F1} [\%] & \textbf{AP} [\%] \\
\hline
0.3 & 2665 & 81 & 22 & 97.05 & 99.18 & 98.10 & 98.96 \\
0.5 & 2645 & 101 & 42 & 96.32 & 98.43 & 97.36 & 97.93 \\
0.7 & 2473 & 273 & 214 & 90.05 & 92.03 & 91.03 & 89.56 \\
\hline
\end{tabular}
\caption{\label{tab:performance_panel_campania} Precision, Recall, F1-score and AP of panel detector for different IoU  on test set of Plant\_Campania. The absolute number of TP, FP and FN is also reported.}
\end{table}

\subsubsection{Defect detector and multi-stage model}
\begin{figure}[!tb]
\centering
	\vspace{0cm}
	\includegraphics[width= \linewidth, keepaspectratio]{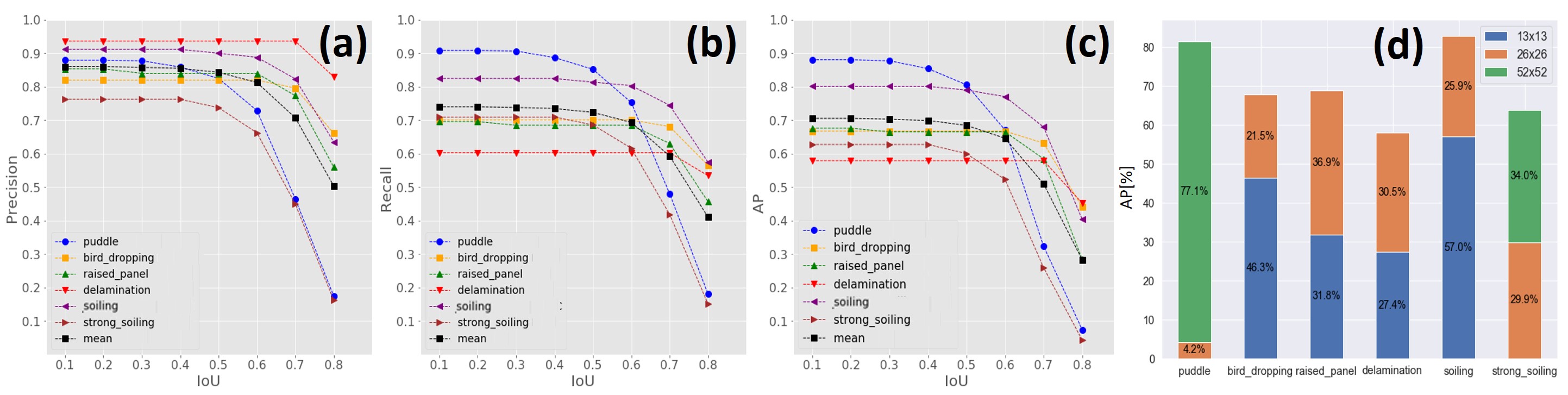}
	\caption{(a) Precision, (b) Recall and (c) AP of defect detector as a function of IoU; (d) AP@0.5 as a function of defect class grouped by YOLOv3 output scale. The PV system under consideration is Plant\_Campania.} 
	\label{fig:defect_detector_performance_campania}
\end{figure}

In Fig.~\ref{fig:defect_detector_performance_campania}(a)-(c) the PREC, REC and AP are shown as a function of IoU, whereas the numerical values @0.5 are reported in Tab.~\ref{tab:performance_defect_campania}.
Metrics for classes bird\_dropping, raised\_panel, delamination, and soiling are flat until IoU= 0.6, in contrast with puddle and strong\_soiling for which degradation starts @0.4 as a consequence of the modest size $<$ 0.4\% (Fig.~\ref{fig:eda_defects_2D_campania}) and the ground truths annotated singularly for each instance. 
The same reason, as well as the different statistics (Tab.~\ref{tab:statistics_campania_dataset_class}), may explain the gap in REC@0.5 (PREC@0.5) of almost 12.8\% (16.2\%) between similar classes soiling and strong\_soiling. 
In general, the larger the statistics available, the higher the recall achievable. 
The detector is most sensitive to class puddle, with a REC@0.5 of almost 85.2\%, and less sensitive (60.3\%) but more precise (93.6\%) for class delamination. Handling unbalancing may help to improve performances for minority classes. On average, an IoU of 0.5 is a satisfactory balance point with a mAP roughly equal to 68.5\% and a mean F1-score of 77.5\%. 

We found also, as expected due to the different defect size already discussed, that accumulation of bird drop and soiling, as well as raised panel and delamination, are roughly equally predicted by output feature maps 13x13 and 26x26, whereas the contribution of the output scale 52x52 becomes relevant for strong soiling deposition and predominant for puddle instances (Fig.~\ref{fig:defect_detector_performance_campania}(d)). In the perspective of a possible integration on-board of the architecture and depending on the more interesting defects to monitor, YOLOv3 can be therefore simplified removing the useless output scales.

\begin{table}[h]
\small
\centering
\begin{tabular}{c|c|c|c|c|c|c|c|c}
\hline
\rowcolor[gray]{0.9} \textbf{class} & \textbf{TP} & \textbf{FP} & \textbf{FN} & \textbf{PREC} & \textbf{REC} & \textbf{F1}& \textbf{AP} &  \textbf{mAP} \\
\rowcolor[gray]{0.9}  & & & & [\%] & [\%] & [\%] & [\%] &  [\%] \\
\hline
puddle(p) & 438  & 93  & 76  & 82.48  & 85.21 & 83.82 & 80.57 & \multirow{6}{*}{68.45} \\
bird\_dropping(bd) & 382  & 84  & 163  & 81.97  & 70.09 & 75.56 & 66.72 &  \\
raised\_panel(rp) & 63  & 12  & 29  & 84.00  & 68.47 & 75.44 & 66.51  & \\
delamination(d) & 44  & 3  & 29  & 93.61  & 60.27 & 73.33 & 57.91 & \\
strong\_soiling(ss) & 59  & 21  & 27  & 73.75  & 68.60 & 71.08 & 60.01 &  \\
soiling(s) & 153  & 17  & 35  & 90.00  & 81.38 & 85.47 & 78.99  & \\
\hline
\end{tabular}
\caption{\label{tab:performance_defect_campania} Evaluation metrics @0.5 on test set of defect detector for Plant\_Campania. The absolute number of TP, FP and FN is also reported.}
\end{table}
\begin{figure}[!tb]
\centering
	\vspace{0cm}
	\includegraphics[width= \linewidth, keepaspectratio]{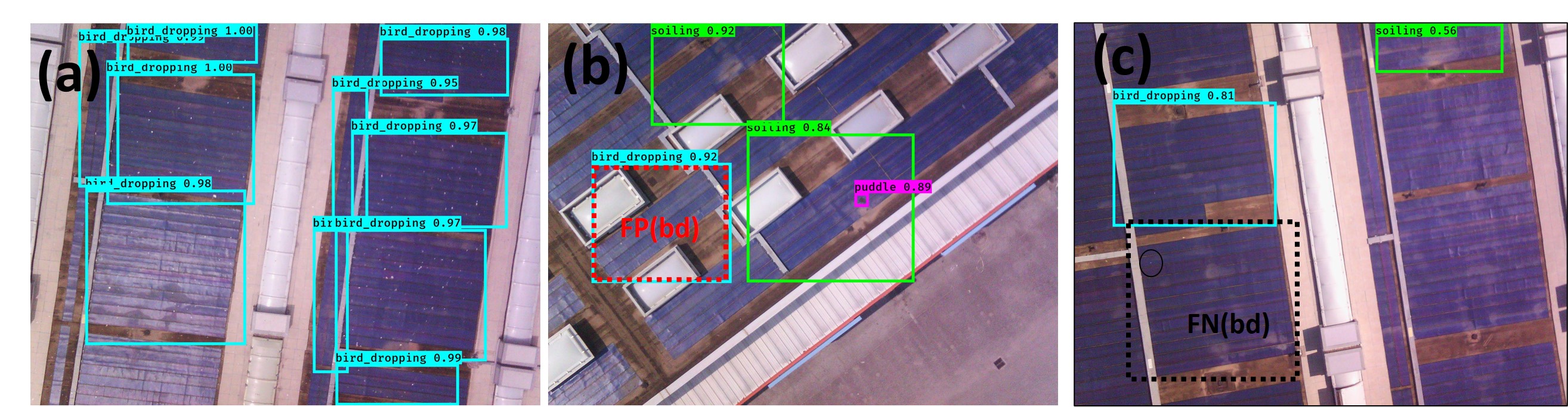}
	\caption{(a) TP (sky blue boxes), (b) FP (highlighted in red) and (c) FN (highlighted in black) for class bird\_dropping (bd) of Plant\_Campania.}
	\label{fig:bird_dropping_detection_campania}
\end{figure} 

\begin{figure}[!tb]
\centering
	\vspace{0cm}
	\includegraphics[width= \linewidth, keepaspectratio]{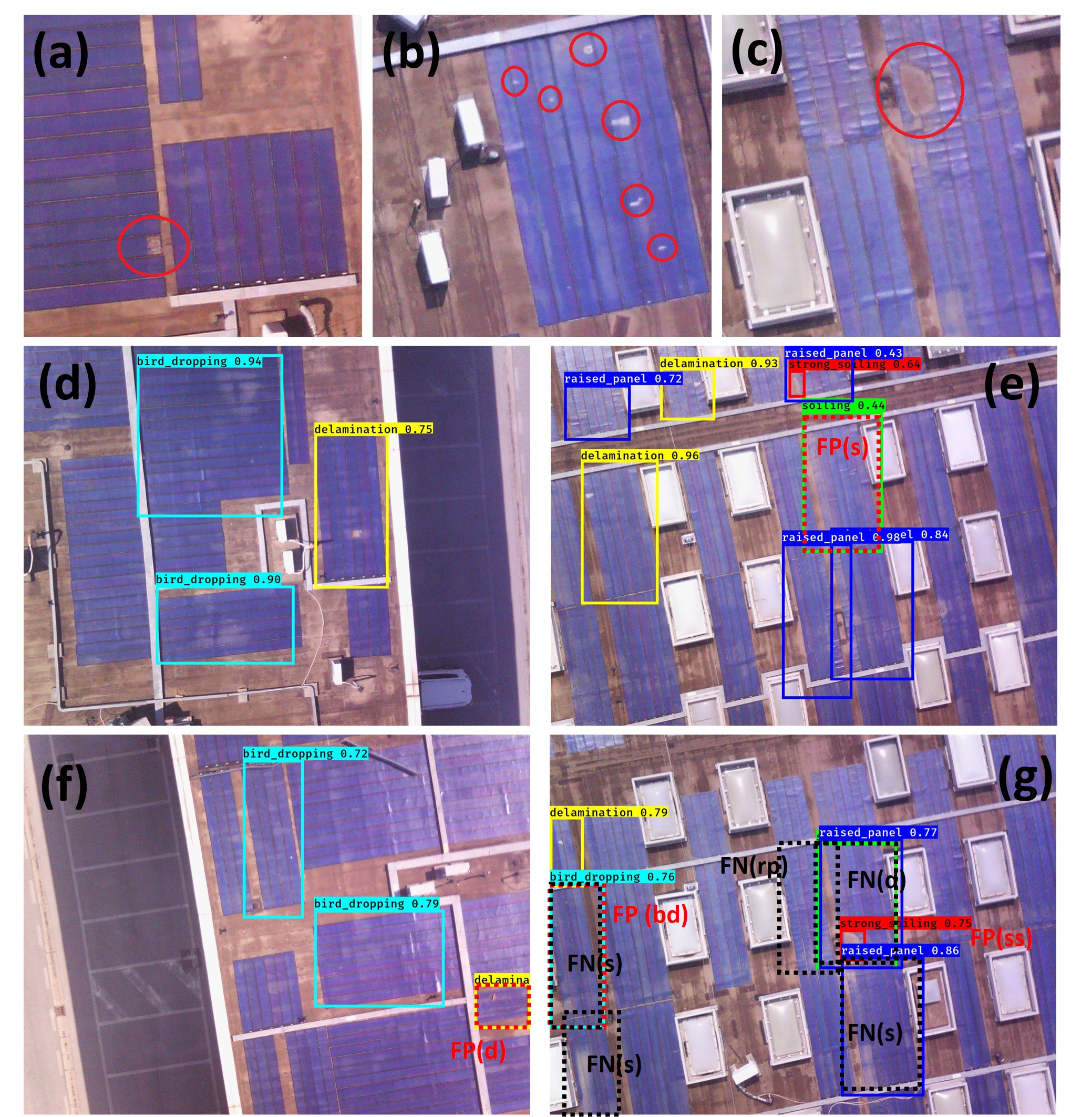}
	\caption{(a)-(c): some occurrences of delamination; (d)-(e) TP (yellow boxes), (f) FP (highlighted in red) and (g) FN (highlighted in black) for class delamination (d) of Plant\_Campania. FP and FN of classes soiling, bird\_dropping and raised\_panel are also included.}
	\label{fig:delamination_detection_campania}
\end{figure} 
\begin{figure}[!tb]
\centering
	\vspace{0cm}
	\includegraphics[width= \linewidth, keepaspectratio]{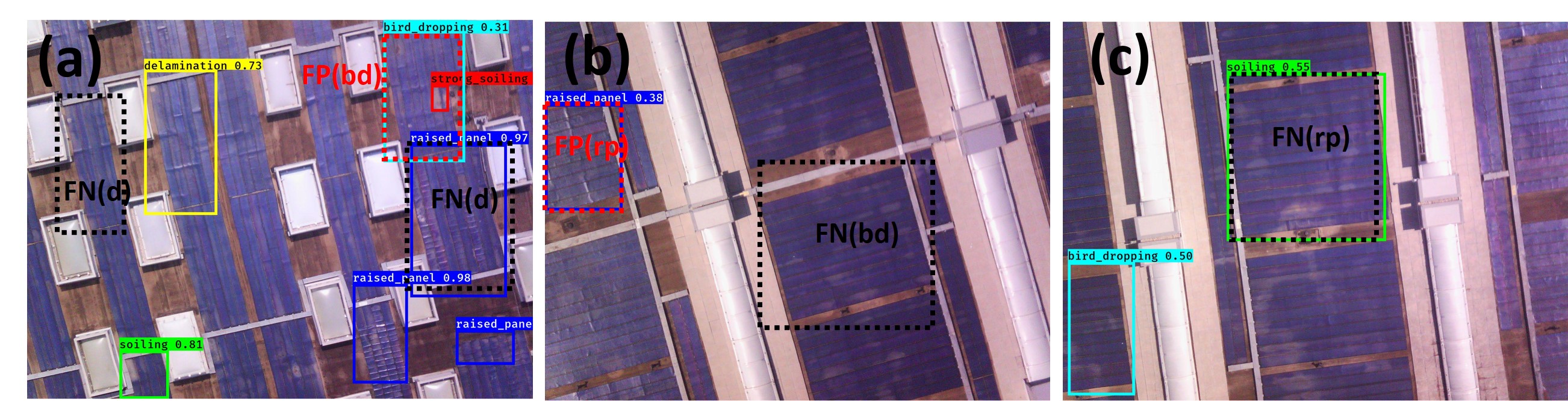}
	\caption{(a) TP (blue boxes), (b) FP (highlighted in red) and (c) FN (highlighted in black) for class raised\_panel of Plant\_Campania. FP and FN of classes bird\_dropping and delamination are also included.}
	\label{fig:raised_panel_detection_campania}
\end{figure} 

\begin{figure}[!tb]
\centering
	\vspace{0cm}
	\includegraphics[width= \linewidth, keepaspectratio]{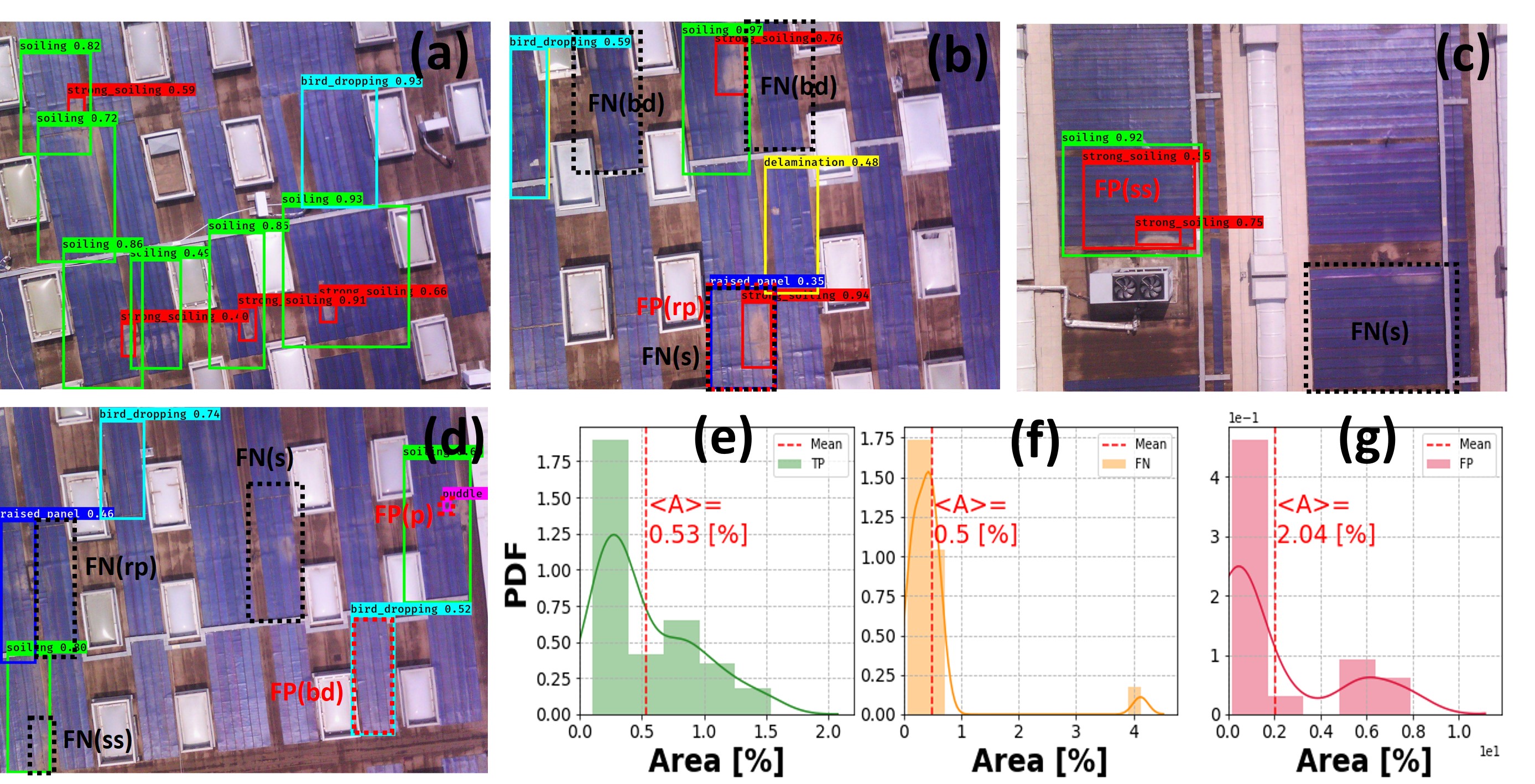}
	\caption{(a)-(b) TP (red boxes), (c) FP (multiple detections of the same instance) and (d) FN (highlighted in black) for class strong\_soiling of Plant\_Campania. FP and FN of classes soiling, bird\_dropping, raised\_panel and puddle are also included. (e)-(g): PDF of the normalized area of TP (e), FN (f) and FP (g) instances. The mean area value is also superimposed for convenience as a red dashed line.}
	\label{fig:strong_soiling_detection_campania}
\end{figure}

\begin{figure}[!tb]
\centering
	\vspace{0cm}
	\includegraphics[width= \linewidth, keepaspectratio]{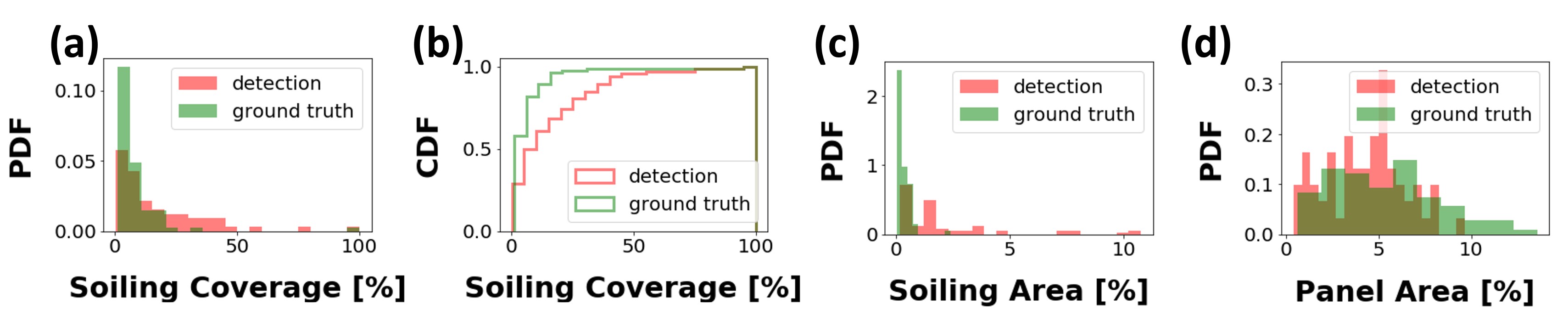}
	\caption{(a)-(b): PDF (a) and CDF (b) of the predicted (red) and ground truth (green) soiling coverage; (c)-(d): predicted (red) and ground truth (green) normalized soiling area (c) and panel area (d). The PV system under consideration is Plant\_Campania.}
	\label{fig:soiling_percentage_info_campania}
\end{figure} 

\begin{figure}[!tb]
\centering
	\vspace{0cm}
	\includegraphics[width= \linewidth, keepaspectratio]{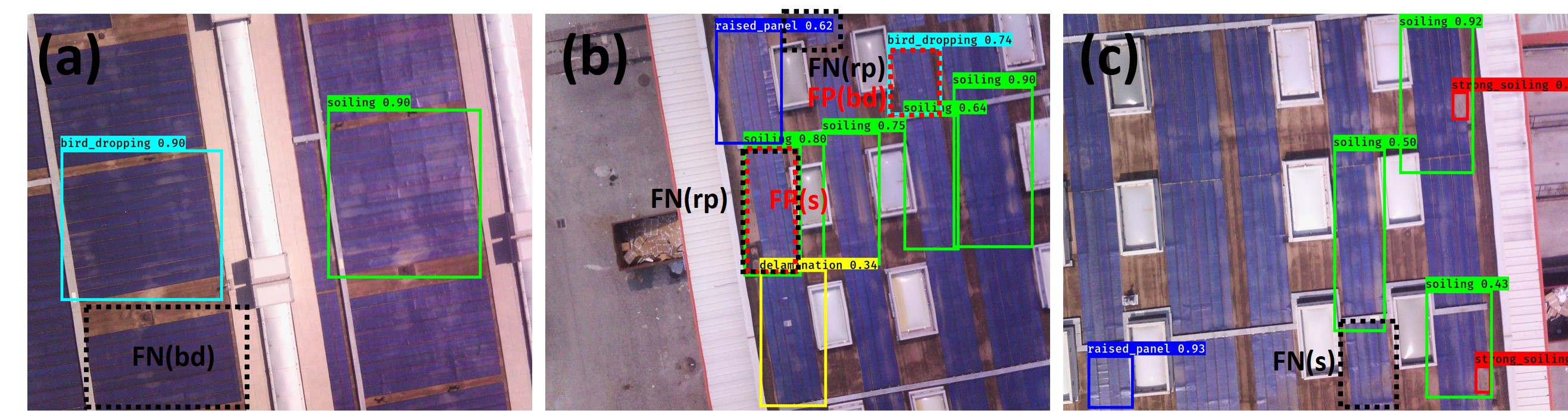}
	\caption{(a) TP (green boxes), (b) FP (highlighted in red) and (c) FN (highlighted in black) for class soiling of Plant\_Campania. FP and FN of classes soiling, bird\_dropping and raised\_panel are also shown.}
	\label{fig:soiling_detection_campania}
\end{figure} 

\begin{figure}[!tb]
\centering
	\vspace{0cm}
	\includegraphics[width= \linewidth, keepaspectratio]{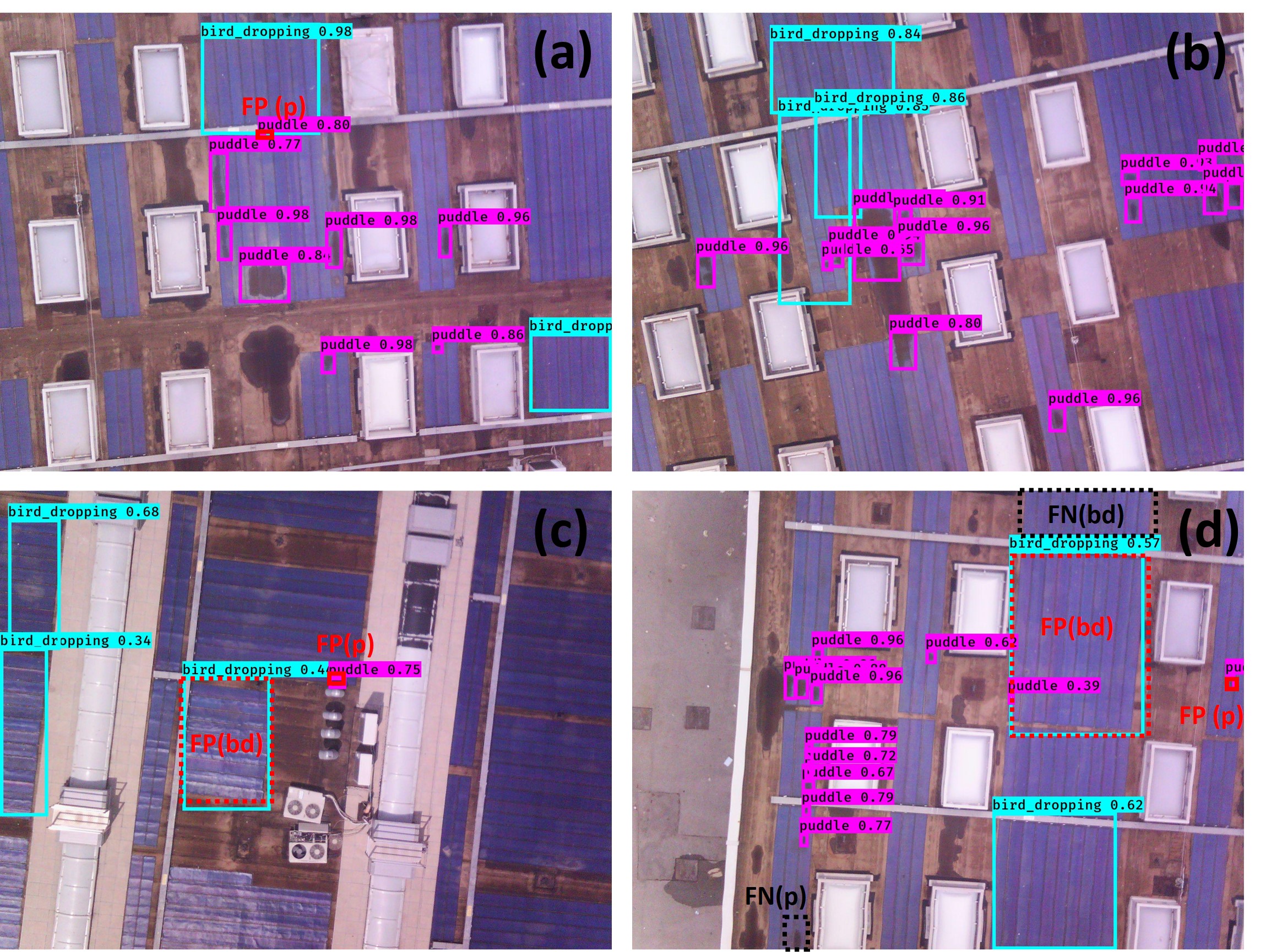}
	\caption{(a)-(b) TP (violet boxes), (c) FP (highlighted in red)  and (d) FN (highlighted in black) for class puddle of Plant\_Campania. FP and FN of class bird\_dropping are also included.}
	\label{fig:puddle_detection_campania}
\end{figure} 

The model is effective in detecting sparse accumulation of bird drop (Fig.~\ref{fig:bird_dropping_detection_campania}a), whereas misdetections appear often in correspondence of panels affected by early delamination on single PV cells (Fig.~\ref{fig:bird_dropping_detection_campania}b), or dust deposition (Figures~\ref{fig:delamination_detection_campania}g,~\ref{fig:raised_panel_detection_campania}a). Missed detections occur instead for very incipient deposition of bird drop, when the impact on power loss should be however negligible (Figures~\ref{fig:bird_dropping_detection_campania}c,~\ref{fig:raised_panel_detection_campania}b,~\ref{fig:strong_soiling_detection_campania}b, ~\ref{fig:puddle_detection_campania}d).

Class delamination has few instances corresponding roughly to 5.3\% of the overall dataset (Tab.~\ref{tab:statistics_campania_dataset_class}). More critically, such instances can differ in shape and colour. Indeed they can be in an advanced status, affecting the entire cell and being visible as a square spot (Fig.~\ref{fig:delamination_detection_campania}a), or also extending over multiple cells (Fig.~\ref{fig:delamination_detection_campania}c). Delamination may be also at an early stage, characterized by an irregular shape (Fig.~\ref{fig:delamination_detection_campania}b). The color may be variable as well, changing from grey (Fig.~\ref{fig:delamination_detection_campania}b) to rusted once the degradation becomes more severe (Fig.~\ref{fig:delamination_detection_campania}a,c). While the model can detect anomalies of all the mentioned types (Fig.~\ref{fig:delamination_detection_campania}d-e), such heterogeneity can justify the not excellent REC@0.5 of almost 60.3\%, due in some circumstances to misdetections as strong\_soiling because of similar colour and shape (Fig.~\ref{fig:delamination_detection_campania}g), or simultaneous occurrence of other anomalies (e.g. raised\_panel, Fig.~\ref{fig:raised_panel_detection_campania}a) which degrades the quality of meaningful features extracted by YOLOv3. Nevertheless, the precision is close to 94\% with very few FPs (Fig.~\ref{fig:delamination_detection_campania}f).  

Concerning raised panels, model missed more often panels affected by concurrent defects including soiling (Fig.~\ref{fig:raised_panel_detection_campania}c), that was recognized as the main cause of degradation by the detector, or incipient raising (Figures~\ref{fig:delamination_detection_campania}g,~\ref{fig:strong_soiling_detection_campania}d). The model is also sometimes confused by sun's glare which produces some bright ripples (Figures~\ref{fig:raised_panel_detection_campania}b,~\ref{fig:strong_soiling_detection_campania}b), causing a misdetection rate 1-PREC of almost 16\% @0.5.

Class strong\_soiling represents the lowest trade-off between PREC and REC with a F1-score of roughly 71.1\%. YOLOv3 is sensitive to such defects until the deposition of dust yields spots with a well delimited pattern (Fig.~\ref{fig:strong_soiling_detection_campania}a-b), whereas it fails  when the defect contour is smooth (Fig.~\ref{fig:strong_soiling_detection_campania}d). It has been found also that false positives have a mean area of almost 2\% of the whole image (Fig.~\ref{fig:strong_soiling_detection_campania}g), i.e. roughly four times larger than true positives (Fig.~\ref{fig:strong_soiling_detection_campania}e), and are more likely to appear as multiple detections of the same defect not suppressed by the NMS algorithm due to modest overlapping (Fig.~\ref{fig:strong_soiling_detection_campania}c).
We verified also that the model tends still to overestimate the soiling deposition (Fig.~\ref{fig:soiling_percentage_info_campania}a), causing a longer tail of the respective estimated PDF with respect to the ground truth one and a consequent underestimation of the Cumulative Distribution Function (CDF), quantifiable in a Kolmogorov-Smirnov index~\cite{Espinar 09} exceeding 0.2 (Fig.~\ref{fig:soiling_percentage_info_campania}b).
This behaviour is mainly due to an over-forecast of the soiling area (Fig.~\ref{fig:soiling_percentage_info_campania}c), rather than an under-forecast of the panel area (Fig.~\ref{fig:soiling_percentage_info_campania}d).

Class soiling has the highest F1-score @0.5 of almost 85.5\%: misdetections corresponds usually at dust deposition lower than expected (Figures~\ref{fig:soiling_detection_campania}b,~\ref{fig:delamination_detection_campania}e), whereas we found that missed detections occur more often for elongated panel stripes (Figures~\ref{fig:soiling_detection_campania}c,~\ref{fig:delamination_detection_campania}g,~\ref{fig:strong_soiling_detection_campania}b-d), since the training statistics was poor as the drone flight on such installations was conducted after it rained. 

It is worth noticing that, unrespective of their smallest size (Fig.~\ref{fig:eda_defects_2D_campania}), YOLOv3 performs well also for class puddle with the best AP@0.5 of about 80.6\%, detecting occurrences with different dimension and shape (Fig.~\ref{fig:puddle_detection_campania}a-b), as a consequence of the multi-scale detection by the FPN. Notably, YOLOv3 learns correctly also the context around the puddle thanks to large receptive field of the network achieved by downsampling layers. Indeed we found that only 4 misdetections out of 93 are located outside the panel area, whereas some others are produced by shadowing of neighbour rooftop installations (Fig.~\ref{fig:puddle_detection_campania}c).

The false alarm filter improves further the precision of the overall model removing the four misdetections of class puddle mentioned above, and without impacting on performances of the other classes. The achieved balanced precision (83.1\%) and recall (85.2\%) outlines that the multi-stage detector can discriminate well the artifacts from real puddle instances (Tab.~\ref{tab:performance_multistage_campania}).

 
\begin{table}[!h]
\small
\centering
\begin{tabular}{c|c|c|c|c|c|c|c}
\hline
\rowcolor[gray]{0.9} \textbf{method} & \textbf{TP} & \textbf{FP} & \textbf{FN} & \textbf{PREC} & \textbf{REC} & \textbf{F1}& \textbf{AP} \\
\rowcolor[gray]{0.9}  & & & & [\%] & [\%] & [\%] & [\%] \\
\hline
defect & 438 & 93 & 76 & 82.48 & 85.21 & 83.82 & 80.57 \\
\rowcolor[rgb]{ .886,  .937,  .855} multi-stage & 438 & 89 & 76 &  \textbf{83.11} & 85.21 &  \textbf{84.14} &  \textbf{80.59} \\
\hline
\end{tabular}
\caption{\label{tab:performance_multistage_campania} Evaluation metrics @0.5 on test set for class puddle of Plant\_Campania. The absolute number of TP, FP and FN is also reported. Results for both defect detector and multi-stage model are presented.}
\end{table}


\section{Conclusion} \label{Conclusion}

Driven by the request of lowering the cost of O\&M in order to maximize plant revenue and by rapidly evolving enabling technologies, most notably UAV devices and artificial intelligence, automatic detection of anomalies in PV panels based on deep learning algorithms is now becoming a hot research topic. This work proposes one of the first UAV-based inspection system based on a multi-stage architecture built on top of the YOLOv3 network. Some of its distinctive key-features include its applicability to both thermal and visible images with very modest customization required and its "Plug and Play" nature, i.e. its fast portability on a park of PV systems of increasing size and different panel technologies. 

To demonstrate its effectiveness, we have presented performances on two large PV plants in the southern of Italy, either on-ground or roof-mounted, with a nominal capacity of tens of MW each for detection of hotspots and other anomalies such as soiling, bird dropping, delamination, or potential issues not discussed yet in literature, such as presence of puddles or panels unglued from the base. 
In particular, unlike traditional O\&M on-site inspection, drone surveys held just after rainfall allowed to identify stagnant water (puddles) that may turn into severe soiling deposition once evaporated or cause moisture ingress in the PV module through eventual micro-cracks or delamination points.
In addition, in case of severe soiling, we return the degree of pollution deposition, based on the predicted defect and panel areas, and its location. Concerning hotspots, we also prescript the recommended intervention based on the predicted defect severity, which can save considerable cost of catastrophic failures related to traditional reactive maintenance strategy.

Results are promising and encourage further research on this topic: panels are detected with an outstanding AP@0.5 of 98\% or more for both PV systems, whereas for IR images collected on Plant\_Sicilia (9~MW) hotspots are identified with an AP@0.4 (AP@0.5) of roughly 88.3\% (66.9\%), and in less than 1~s per image. Comparable accuracy was presented in Ref.~\cite{Herraiz 20}, where however a much smaller PV system (100~KW) was discussed and no prescriptive outcomes were provided based on the identified hotspots. 
Concerning instead visible images captured on Plant\_Campania (21~MW), defects are detected with an AP@0.5 ranging from 57.9\% for delamination up to 79.0\% (80.6\%) for soiling (puddle). On average, we achieve a mAP@0.5 of 68.5\%. We found also that the model still tends to over-forecast the percentage panel area affected by soiling, which is essentially due to a positive bias in the predicted soiling area. The multi-stage architecture is also effective in removing artifacts identified by the defect detector on the background.

Nevertheless, some further researches must be still carried out. 
When doing O\&M services above a huge PV plant with an extension of more than 100~ha, image acquisition shall be planned with a cost-effective approach. In this scenario, drone hovering for static image acquisition of a few PV modules~\cite{Herraiz 20} is not sustainable and a continuous acquisition flight plan shall be adopted. In such conditions, final defect localization error is influenced by the following: GNSS vertical and horizontal errors, gimbal pointing error, time-shift between metadata and image registration related to flight speed. To reduce it, we propose in a future work to replace drone standard GPS with more accurate GNSS-RTK receivers (whose typical horizontal error is $\pm~1~cm$ instead of $\pm~1.5~m$), and to pre-process images with orthomosaic techniques, that would impact the time-shift error. The combination of both error reduction strategies should allow a defect localization at cell-level. It would also introduce an immediate benefit for O\&M cost reduction, enabling O\&M operator to avoid time and specialized equipment wasted to repeat on-site inspection analysis with handheld devices with the aim to identify the correct target among adjacent PV panels in the localization error radius. Adoption of orthomosaics, which requires to handle high resolution images, may also introduce further benefits to investigate: the same rotation will be applied at the same time to all the images of a given PV plant, that could lead to a rotation discrepancies reduction. Furthermore RGB and IR mosaics can then be easily overlayed, potentially unveiling new kind of data fusion analysis. 

Also, estimation of power loss of defective modules based on acquisition of electrical tags from PV strings, as well as prediction of the dust distribution on soiled solar modules, which in turn affect their efficiency, may be exploited by plant operator to schedule more intelligent maintenance operations such as module cleaning, repair or replacement.


Robustness can be further improved by extending the inspection service on novel PV sites and implementing a two-stage finetuning, similarly to Ref.~\cite{Miao 19} for automatic inspection of power lines, by preliminarily training on a generic dataset of defective and normal panels and then finetuning on data of the specific PV installation starting from a pre-trained model on the basic dataset. In this way the basic model can be repeatedly exploited without re-executing the first-stage finetuning, consequently reducing the training cost and improving the flexibility of the training process.

Finally, actually the model can post-process automatically batches of images captured on-site and transferred on a standalone computer, whereas real-time processing directly on-board of drone need further research to shrink the YOLO architecture in order to satisfy the constraints imposed by the embedded hardware in terms of size, speed and accuracy. 
Preliminary simulations aimed to verify the impact of YOLOv3's output scales on accuracy unveil a strong coupling with the desired defect type to detect, thus prompting the possibility to streamline the YOLOv3 architecture removing the outputs less efficient for defect detection.  
\section*{Author Contributions}{Conceptualization, A.B.; methodology, A.B. and A.D.T.; software, A.D.T. and A.B; validation,  A.D.T. and A.B.; UAV-based data acquisition, GF; data curation, A.D.T., B.M. and G.F.; writing---original draft preparation, A.B. and A.D.T.; writing---review and editing, A.B. and G.F.; visualization, A.D.T. and A.B.; supervision, A.B.. All authors have read and agreed to the published version of the manuscript.}

\section*{Acknowledgment}
This research did not receive any specific grant from funding agencies in the public, commercial, or
not-for-profit sectors. The authors acknowledge the company i-EM srl (\url{https://www.i-em.eu/}) for support in the data collection stage and the anonymous reviewers for their contribution to improve this paper.

\end{document}